\documentclass[final]{cvpr}

\usepackage{times}
\usepackage{epsfig}
\usepackage{graphicx}
\usepackage{amsmath}
\usepackage{amssymb}
\usepackage{xcolor,colortbl}
\usepackage{subfigure}
\usepackage{caption}
\usepackage{array}
\usepackage{makecell}
\usepackage{bbding}
\definecolor{my_blue}{rgb}{0.58,0.73,0.91}
\definecolor{my_red}{rgb}{0.91,0.63,0.58}

\usepackage[pagebackref=true,breaklinks=true,colorlinks,bookmarks=false]{hyperref}

\def\cvprPaperID{4624} 
\def\confYear{CVPR 2021}

\begin{document}

\title{Learning Depth via Leveraging Semantics: Self-supervised Monocular Depth Estimation with Both Implicit and Explicit Semantic Guidance}

\author{Rui Li, Xiantuo He, Danna Xue, Shaolin Su, Qing Mao, Yu Zhu, Jinqiu Sun, Yanning Zhang\\
Northwestern Polytechnical University\\
{\tt\small lirui.david@gmail.com, sunjinqiu@nwpu.edu.cn}



}

\maketitle

\begin{abstract}
Self-supervised depth estimation has made a great success in learning depth from unlabeled image sequences. While the mappings between image and pixel-wise depth are well-studied in current methods, the correlation between image, depth and scene semantics, however, is less considered. This hinders the network to better understand the real geometry of the scene, since the contextual clues, contribute not only the latent representations of scene depth, but also the straight constraints for depth map. In this paper, we leverage the two benefits by proposing the implicit and explicit semantic guidance for accurate self-supervised depth estimation. We propose a Semantic-aware Spatial Feature Alignment (SSFA) scheme to effectively align implicit semantic features with depth features for scene-aware depth estimation. We also propose a semantic-guided ranking loss to explicitly constrain the estimated depth maps to be consistent with real scene contextual properties. Both semantic label noise and prediction uncertainty is considered to yield reliable depth supervisions. Extensive experimental results show that our method produces high quality depth maps which are consistently superior either on complex scenes or diverse semantic categories, and outperforms the state-of-the-art methods by a significant margin.

\end{abstract}

\section{Introduction}

\begin{figure}[ht]
  \centering
  \subfigure{
     \begin{minipage}[c]{0.01\linewidth}
                \centering
                 \small
                \rotatebox{90}{\makebox{Input}}
      \end{minipage}
      \begin{minipage}[c]{1\linewidth}
          \centering
          \includegraphics[width=0.45\linewidth]{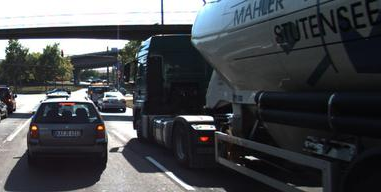}
          \includegraphics[width=0.45\linewidth]{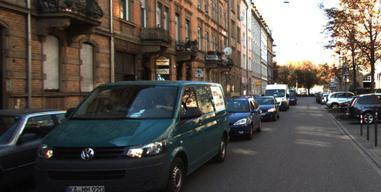}
      \end{minipage}%
  }\vspace{-8pt}
  \subfigure{
      \begin{minipage}[c]{0.01\linewidth}
                \centering
                 \small
                \rotatebox{90}{\makebox{PackNet \cite{guizilini20203d}}}
      \end{minipage}
      \begin{minipage}[c]{1\linewidth}
          \centering
          \includegraphics[width=0.45\linewidth]{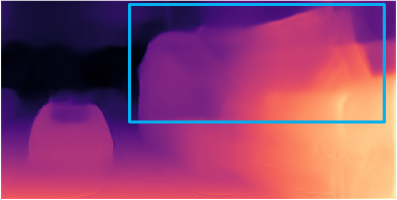}
          \includegraphics[width=0.45\linewidth]{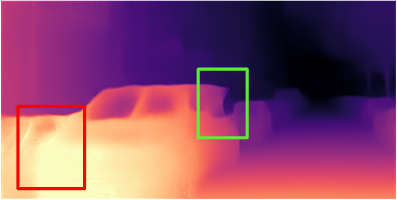}
      \end{minipage}%
  }\vspace{-8pt}
  \subfigure{
     \begin{minipage}[c]{0.01\linewidth}
                \centering
                 \small
                \rotatebox{90}{\makebox{\textbf{Ours}}}
      \end{minipage}
      \begin{minipage}[c]{1\linewidth}
          \centering
          \includegraphics[width=0.45\linewidth]{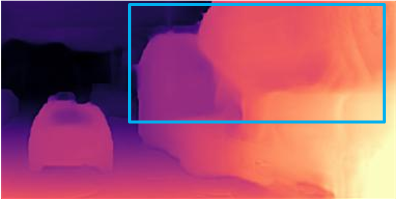}
          \includegraphics[width=0.45\linewidth]{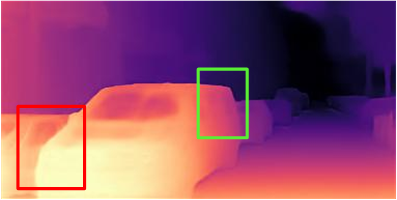}
      \end{minipage}%
  }\vspace{-8pt}
     
  \centering
  \caption{\textbf{Example of monocular depth estimation}. Compared with the SOTA method \cite{guizilini20203d}, our method successfully recognizes the object depth from unusual categories (left blue box). At the mean time, our method generates high quality depth map which possesses smooth depth change inside the object region (right green box) and sharp depth edge across object borders (right red box).}
  \label{fig:teaser}
  \vspace{-15pt}
  \end{figure}

Depth estimation is a long-standing problem in computer vision, which is widely used in robotic perception, autonomous driving as well as multimedia applications \cite{viereck2017learning,Gaidon:Virtual:CVPR2016}, \etc. Compared with classical geometry-based methods \cite{schonberger2016structure,li2019robust} which estimate depth using stereo or sequential images, learning-based methods \cite{eigen2014depth,eigen2015predicting,li2015depth} are able to conduct pixel-wise dense predictions given only a single image as input. However, as the deep neural network requires \textcolor{black}{a} large amount of labeled data for training, ideal depth labels are hard to acquire due to the expensive LiDAR sensors as well as the sparse ground truth annotations. Under these circumstances, self-supervised depth estimation \cite{godard2017unsupervised,zhou2017unsupervised,mahjourian2018unsupervised,yin2018geonet,godard2019digging,guizilini20203d,li2020enhancing} has become a new trend that it managed to generate high quality depth maps using only image sequences for supervision.
\par

\textcolor{black}{
Despite the impressive results, most current methods use only image representation to model image depth, and seldom consider how the semantic information can be utilized. Image semantics are, however, highly coupled with image representation and scene depth. Therefore it's necessary to introduce image semantics to assist depth estimation.
}

\textcolor{black}{
We address the benefits of leveraging image semantic guidance in two aspects: on the one hand, it's highly ill-posed to directly estimate scene depth from images skipping object category information \cite{guizilini2020semantically}. Specifically, the appeared object size and location in the image, which are commonly regarded as direct depth indicators \cite{guizilini2020semantically,dijk2019neural}, do not straightly correlate with actual image depth. As shown in the left column of Figure \ref{fig:teaser}, SOTA method \cite{guizilini20203d} failed to infer the depth of a truck with unusual vehicle properties than other cars.
On the other hand, image semantics also provide straight clues for object depth. Typically, the semantic borders also correspond to depth borders, and the depth distribution of each individual object is category-specific. For example, objects such as ``person'' and ``traffic sign'' possess uniform depth in the scene, while categories ``buildings'' and ``road'' exhibit gradual depth change. This raises new challenges for conventional methods. As shown in the right column, \cite{guizilini20203d} produces obvious ``bleeding artifacts'' \cite{zhu2020edge} in the depth borders, and the abnormal sudden depth changes are also observed inside an continuous object area.
}


\textcolor{black}{
In order to fully exploit the benefits of image semantics for depth estimation, we propose to impose semantic guidances in \textit{implicit} and \textit{explicit} ways, respectively. 
In our framework, we extract semantic features by a semantic segmentation branch, and fuse them with a depth estimation branch at feature level to assist model prediction. In this way, image semantics are combined with image representation to \emph{implicitly} guide depth estimation. 
Based on the analysis that the depth distribution is highly related to the semantic categories, we further propose the Semantic-aware Spatial Feature Alignment (SSFA) module, which spatially aligns depth and semantic features via constraining depth distribution to be consistent inside the same category area, while to be different across other categories. During this process, semantic category information is used as external prior to guide the feature alignment.
We further propose the semantic-guided ranking loss to \emph{explicitly} improve the estimated depth map quality. Given semantic labels, we sample a set of cross-edge point quadruplets to explicitly constrain the depth map to be smooth inside the semantic object areas, while to have sharp depth edges across the semantic borders. Different from other edge-based methods \cite{zhu2020edge,chen2019towards}, we take the impact of erroneous semantic labels into consideration, and propose robust point pair sampling strategy and semantically uncertainty-aware weighting to alleviate semantic label noise.
As shown in Figure \ref{fig:teaser}, the proposed implicit and explicit semantic guidances yield accurate depth estimation under different challenging scenarios.
}

\par
Our contribution can be concluded as follows:
\begin{itemize}
\item 
We address the advantages of leveraging image semantics for both image depth inference and depth map fining. We thus propose \emph{implicit} and \emph{explicit} semantic guidances to fulfill the two objectives respectively.

\item 
We propose a novel Semantic-aware Spatial Feature Alignment (SSFA) scheme to effectively guide depth estimation in an implicit manner, and propose the semantic-guided ranking loss to explicitly improve the accuracy of estimated depth map.

\item 
Our method generates consistently superior results across different scenarios and semantic categories, quantitative results on KITTI show that it outperforms the state-of-the-art self-supervised monocular trained methods by a significant margin.
\end{itemize}

\begin{figure*}
\centering
\includegraphics[width=0.95\linewidth]{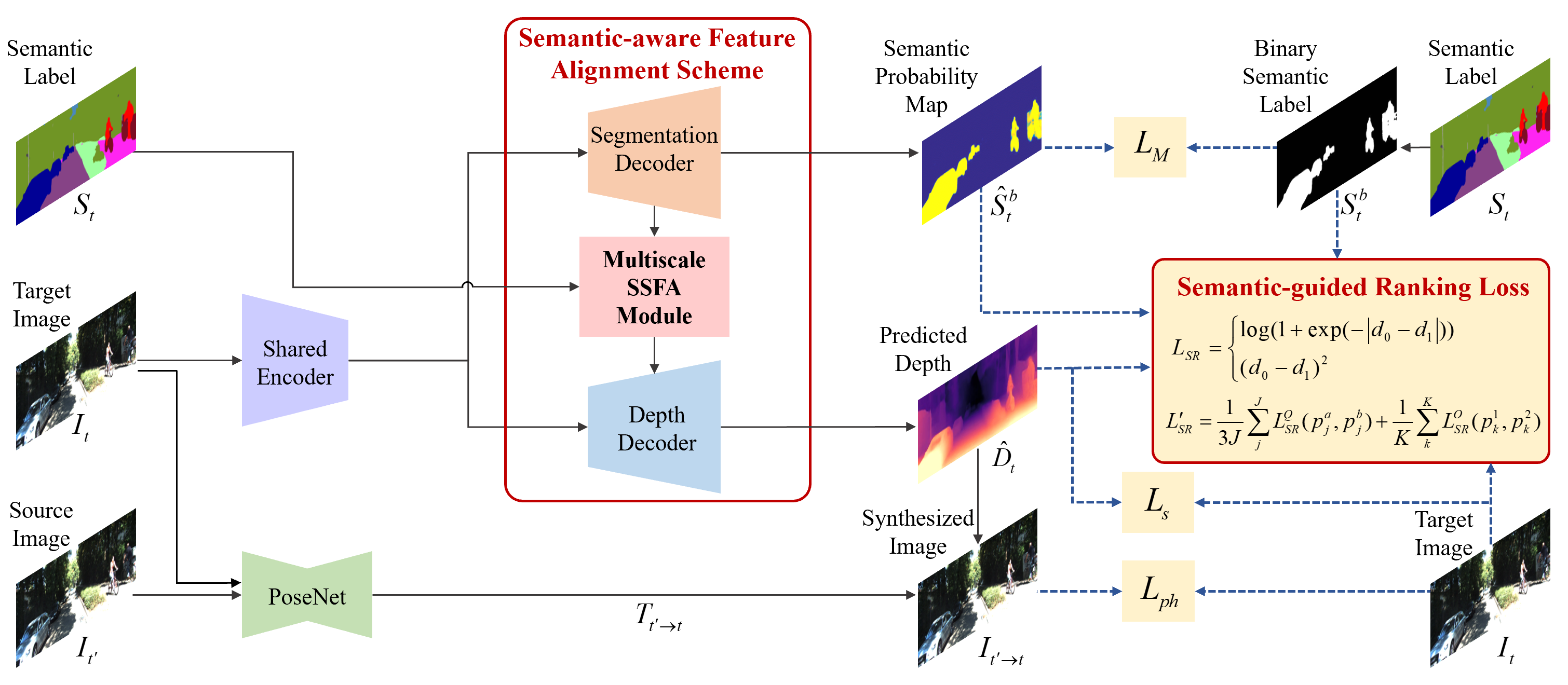}
\centering
\caption{\textbf{The overview of the proposed architecture}. We follow the basic self-supervised depth estimation framework and extend it with both implicit and explicit semantic guidance. The SSFA scheme is proposed to implicitly make depth network semantically aware, while the proposed semantic-guided ranking loss constrains the depth to be consistent with semantic contextual.}
\label{fig:overview}
\vspace{-10pt}
\end{figure*}

\section{Related Work}\label{sec:related_work}
\textbf{Self-supervised depth estimation.} Self-supervised methods cast the depth supervision problem into image-based supervision, which enables learning without depth annotations. As the pioneer methods \cite{godard2017unsupervised,garg2016unsupervised} train depth networks using stereo image pairs, Zhou \etal \cite{zhou2017unsupervised} propose a more generalized pipeline, which enables training with pure image sequences. After that, great progress has been made to improve self-supervised framework in terms of loss, occlusion removal as well as architectures. Yang \etal \cite{yang2020d3vo:} and Li \etal \cite{li2020enhancing} enhance the photometric loss to be robust towards illumination variance, while Shu \etal \cite{shu2020feature} propose the feature-metric loss which improves the loss back propagation on low gradient areas. In order to solve the scene occlusion as well as object motion issues during depth training, several methods \cite{zhou2017unsupervised,godard2019digging,bian2019unsupervised,wang2019unsupervised} propose both learning-based and geometric selective mask that filter out the unreliable losses.
\textcolor{black}{In order to exploit more information for self-surpervised methods} , optical flow is introduced \cite{ranjan2019competitive,wang2019unos,yin2018geonet} for extra constraints. Pseudo depth is also leveraged as extra prior information \cite{zhao2020towards}. In terms of new self-supervised architectures, Guizilini \etal \cite{guizilini20203d} propose novel packing and unpacking modules which preserves more detailed depth predictions. However, in this paper, \emph{for fair comparison, we rule out the influence of new network architectures when comparing with the state-of-the-arts}.
\par
\textbf{Semantic guidance for depth estimation.} Semantic segmentation \textcolor{black}{has shown} its effectiveness for depth estimation in previous works \cite{choi2020safenet,guizilini2020semantically,chen2019towards,ramirez2018geometry}. The methods can be categorized into two groups 
\textcolor{black}{according to how image semantics are used}.
The first group of methods offer implicit semantic guidance via providing feature-level information. Chen \etal \cite{chen2019towards} generate both depth and semantic maps with a unified scene representation. Guizilini \etal \cite{guizilini2020semantically} propose to feed the PAC enhanced \cite{su2019pixel} pre-trained segmentation features for depth estimation. Choi \etal \cite{choi2020safenet} leverage the semantic network and feed the features with cross-propagation and affine-propagation unit. \textcolor{black}{Though we also fuse image depth with semantics in feature level, we align semantic features in a more solid way through our proposed SSFA module, which provides image semantic guidance with distinctly more fidelity and persistence.}

\par
The other group of methods use semantic categories to explicitly constrain or supervise depth networks. Ramirez \etal \cite{ramirez2018geometry} and Chen \etal \cite{chen2019towards} constrain the smoothness of the depth map using semantic maps. Casser \etal \cite{casser2019depth} and Klingner \etal \cite{klingner2020self} handle the dynamic moving object issue via specifying moving areas with semantic object labels, Zhu \etal \cite{zhu2020edge} proposed a stereo-based method which leverages the binary semantic map to generate the edge-aligned pseudo depth labels for direct supervision. Wang \etal \cite{wang2020sdc} propose to learn semantic category-specific depth by a divide-and-conquer strategy. 
\textcolor{black}{Compared with previous explicit semantic constrains, the proposed semantic-guided ranking loss further takes semantic label noise and prediction uncertainty into account, thus lead to more accurate and reliable depth supervision.}

\section{The Proposed Method} \label{sec:method}
Self-supervised depth estimation usually takes image triplet $(I_{t-1},I_{t},I_{t+1})$ as input, where $I_{t}$ is the target image and $I_{t^{'}} \in \{I_{t-1},I_{t+1}\}$ belongs to the source images. During training, $I_{t}$ is fed into the depth network $f_{D}$ to get the predicted depth $\hat{D}_{t} \in \mathbb{R}^{H \times W}$, where $H$ and $W$ are the image height and width. $(I_{t},I_{t^{'}})$ are put into the motion network to get the relative motion $T_{t^{'} \rightarrow t}$. Then, the synthesized image $I_{t^{'} \rightarrow t}$ can be computed with $I_{t}$, $\hat{D}_{t}$ and $T_{t^{'} \rightarrow t}$ \cite{zhou2017unsupervised}. The depth network is trained by back-propagating the photometric loss between the synthesized images $I_{t^{'} \rightarrow t}$ and the target image $I_{t}$, as proposed by Godard \etal \cite{godard2019digging}
\begin{equation}
\begin{aligned}
L_{ph}(I_{t},I_{t^{'} \rightarrow t})=\min_{t^{\prime}}(\frac{\alpha}{2}(1-\operatorname{SSIM}\left(I_{t}, I_{t^{\prime} \rightarrow t}\right)) \\ +(1-\alpha)\left\|I_{t}-I_{t^{\prime} \rightarrow t}\right\|_{1} ),
\end{aligned}
\end{equation}
where SSIM denotes the structural similarity index \cite{wang2004image}, $\alpha$ refers to the weighting factor which is commonly set to $0.85$ \cite{godard2019digging,godard2017unsupervised,zhou2017unsupervised}. The first-order depth smoothness loss $L_{s}$ is also set with the weighting factor of $10^{-3}$ \cite{godard2017unsupervised}.
\par
\textcolor{black}{
In this paper, we follow the basic self-supervised framework but extend it with implicit and explicit semantic guidances for accurate depth estimation. During depth inferring, we introduce a SSFA module to incorporate image semantics, which implicitly makes the estimated depth semantically aware. We further explicitly constrain the estimated depth map with semantic-guided ranking loss, and improve the accuracy and reliability of our predicted depth map. The overview of our model is shown in Figure \ref{fig:overview}.
}

\subsection{Semantic-aware Feature Alignment Scheme}
We enhance the depth estimation performance implicitly by providing semantic feature representations to the depth network. Thus, a semantic branch is proposed to offer semantic category-level information in a multi-scale scheme, see Section \ref{sec:multitask}. Consider the semantic and depth features are from the different task domains, we align the features by the proposed semantic-aware spatial feature alignment (SSFA) module, \textcolor{black}{
utilizes external semantic category-level labels for better semantic guidance, see Section \ref{sec:SSFA}.
}

\subsubsection{Semantic Multi-task Scheme}\label{sec:multitask}
\begin{figure}
\centering
\includegraphics[width=1.0\linewidth]{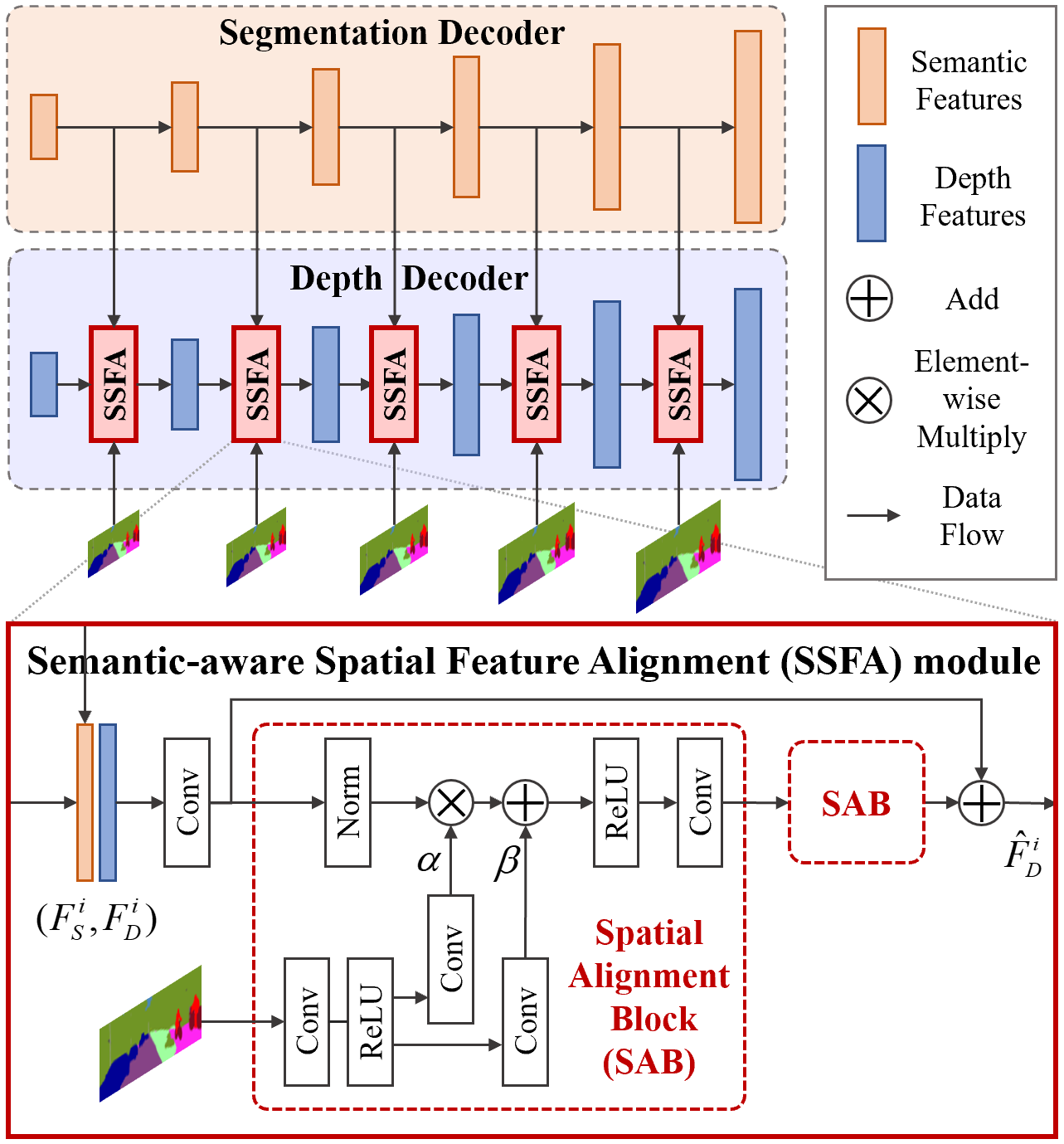}
\centering
\caption{\textbf{The proposed semantic-aware feature alignment scheme}. We propose an extra semantic branch to implicitly offer semantic information to the depth network in a multi-scale manner. During the process, a semantic-aware spatial alignment (SSFA) module is proposed which improves the fusion of depth and semantic features via conducting spatial operations learned from the semantic guidance.
}
\label{fig:SSFA}
\vspace{-10pt}
\end{figure}

We extend the general self-supervised depth estimation framework by add an extra semantic branch $f_{S}$, which shares the same encoder with the depth network, as shown in the top of Figure \ref{fig:SSFA}. Given the input image $I$, a semantic probability map $\hat{S}^{b} \in [0,1]^{H\times W}$ is generated and supervised by the given binary semantic label $S^{b} \in \{0,1\}^{H\times W}$
\begin{equation}
   L_{M}=BCELoss(\hat{S}^{b},S^{b}),
\end{equation}
where $BCELoss(\cdot)$ denotes the binary cross-entropy loss. The semantic labels can be either groundtruth or pre-computed labels from other methods, we use the latter \cite{zhu2019improving} in our experiments due to the lack of semantic groundtruth annotations. 
Let $S \in \mathbb{M}^{H\times W}$ the full semantic label, where $\mathbb{M}$ is the integers denoting the semantic categories. We generate the binary semantic label $S^{b}$ by specifying foreground objects of the given $S$.
The reasons for generating the binary semantic map are twofold, 1) it provides comparable informative semantic features to the depth branch as the full semantic map does \cite{anonymous2021semanticguided}, 2) when using pre-computed semantic labels, cross-domain trained segmentation method \cite{zhu2019improving} achieves much higher mIoU for binary predictions than full category predictions, which indicates that binary labels are more reliable as the input. Experimental details can be found on Section \ref{sec:diff_semdata}.
The semantic branch is trained together with the depth network, and provides latent semantic features to the depth decoding layers in a multi-scale manner, as shown in Figure \ref{fig:SSFA}.

\subsubsection{Semantic-aware Spatial Feature Alignment}\label{sec:SSFA}
Since the semantic and depth features are from different task domains, the simple fusions (direct concatenation or convolution, \etc.) do not fully exploit the potential of image semantics for depth estimation. In this paper, consider the depth distribution of the scene is category-specific, we deduce the distribution of depth features should also be category-specific. Thus, we propose the Semantic-aware Spatial Feature Alignment (SSFA) module, which takes the semantic labels as external prior, to align the features and make them conform to the category-specific distributions, via semantic-aware spatial normalization operations.
\par
As shown in Figure \ref{fig:SSFA}, during network training, we have multi-scale semantic features $F^{i}_{S} \in \mathbb{R}^{C_{i}\times H \times W}$ and depth features $F^{i}_{D} \in \mathbb{R}^{C_{i}\times H \times W} $, where $C_{i}$ denotes the channels of the feature block and $i \in \{0,1,2,3, 4\}$. We propose $5$ SSFA modules corresponding to the number of scales. For each input pair $(F^{i}_{S}, F^{i}_{D})$, the corresponding SSFA module outputs the aligned feature $\hat{F}^{i}_{D}$ as the input of the next scale. 
\par
For each SSFA module, we propose two semantic-aware spatial alignment blocks (SAB) with residual connection as shown in the bottom of Figure \ref{fig:SSFA}. The SSFA module first fuses the two features together and then put the coarsely fused feature $F^{i}_{S+D}$ into the two blocks. Given an input feature $F$ to the spatial alignment block, the block first normalize the input feature in the channel-wise manner, then spatially align the normalized feature $F_{norm}$ with the learned category-specific multiplicative factor $\alpha$ and additive factor $\beta$, as inspired by \cite{park2019semantic,wang2018recovering}
\begin{equation}
   F_{aligned}=\alpha \otimes F_{norm} \oplus \beta,
\end{equation}
where $\otimes$ and $\oplus$ denote element-wise multiplication and addition, $\alpha$ and $\beta$ are learned from the semantic labels via simple conventional layers. After the multi-level processing with the proposed SSFA modules, the output features $\hat{F}^{i}_{D}$ are endowed with better representation capabilities that both semantic and depth features are well aligned by the category-level semantic guidance.

\subsection{Semantic-guided Ranking Loss}
In this section, we explicitly constrain the generated depth map to be consistent inside semantic categories and to be sharp cross the category borders. We resort to the ranking loss \cite{chen2016single} for it directly constrains the depth differences, which is more suitable to constrain depth sharpness/smoothness.
Different from the existing depth ranking loss \cite{chen2016single,xian2020structure} which use the point pair depth similarity to specify the maximize or minimize operation of depth difference, we instead propose a semantic-guided ranking loss, which maximize/minimize the point pair depth difference according to their semantic belongings.
For a sampled point pair $(p_{0},p_{1})$ with depth $(d_{0},d_{1})$, the semantic-guided ranking loss $L_{SR}$ can be formulated as
\begin{equation}\label{eq:ranking_loss}
L_{SR}\left(p_{0},p_{1}\right)=\left\{\begin{array}{ll} \log \left(1+\exp \left(-|d_{0}-d_{1}|\right)\right), & \ell=1  \\ \left(d_{0}-d_{1}\right)^{2}, & \ell=0\end{array}\right.
\end{equation}
where $\ell$ is the semantic belonging indicator denoting the semantic relationship between point pairs

\begin{equation}
\ell=\left\{\begin{array}{ll}1, & S^{b}(p_{0}) \neq S^{b}(p_{1})  \\ 0, & S^{b}(p_{0})=S^{b}(p_{1})\end{array}\right.
\end{equation}
where $S_{b}$ is the binary semantic label. 
If $(p_{0},p_{1})$ lie across the semantic border, $\ell=1$, we constrain the depth difference between $(d_{0},d_{1})$ to be large for sharp depth edges. Otherwise, we minimize the depth difference between $(d_{0},d_{1})$ to make the depth smooth inside the same semantic category. Consider the proposed loss relies explicitly on the semantic borders, little noise on the semantic labels will cause erroneous constrains for the predicted depth. To address this problem, a cross-border point pair sampling strategy (see Section \ref{sec:pts_sampling}) and a semantically uncertainty-aware weighting factor (see Section \ref{sec:uncer_factor}) is proposed which consider the noise of input semantic labels. 
\par

\begin{figure}
\centering
\includegraphics[width=1.0\linewidth]{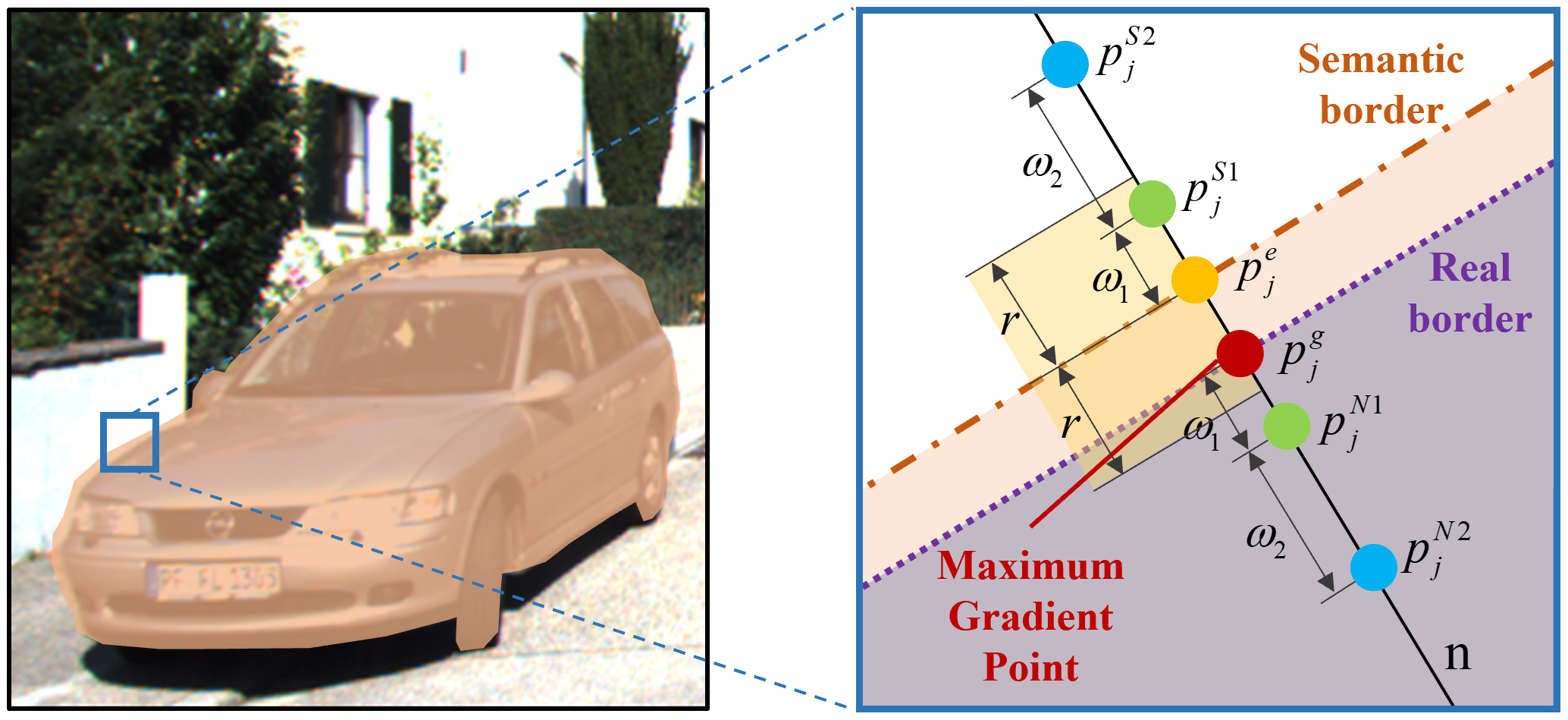}
\centering
\caption{\textbf{The proposed semantic-guided edge sampling strategy}. For every edge point $p_{j}^{e}$, we sample a point quadruplet $Q_{e}=\{p^{S1}_{j},p^{S2}_{j},p^{N1}_{j},p^{N2}_{j}\}$ along orthogonal line $\mathbf{n}$. To find the real border, we search the max gradient point $p_{j}^{g}$ along $\mathbf{n}$ in range $[-r,r]$, and make the cross-border point pair $(p^{S1}_{j},p^{N1}_{j})$ to clamp $p_{j}^{e}$ and $p_{j}^{g}$.
}
\label{fig:edge_sample}
\vspace{-10pt}
\end{figure}

\subsubsection{Cross-border Point Pair Sampling Strategy}\label{sec:pts_sampling}
Inspired by \cite{xian2020structure}, we propose to sample a set of point quadruplets $Q=\{q_{j}\}_{j=1}^{J}$ on the semantic borders to compute the ranking loss, where $J$ is the number of edge points. We first compute the edge map of the binary semantic label, and for every edge point $p_{j}^{e}$, we sample a quadruplet $q_{j}\in Q$, $q_{j}=\{p^{S1}_{j},p^{S2}_{j},p^{N1}_{j},p^{N2}_{j}\}$ which contains $4$ co-linear points lying along the orthogonal line $\mathbf{n}$ crossing the edge point. As shown in Figure \ref{fig:edge_sample}, point $p^{S1}_{j}$ and $p^{S2}_{j}$ lie on one side of $p_{j}^{e}$, then point $p^{N1}_{j}$ and $p^{N2}_{j}$ lie on the other. Totally three point pairs $[(p^{S1}_{j},p^{S2}_{j}),(p^{S1}_{j},p^{N1}_{j}),(p^{N1}_{j},p^{N2}_{j})]$ are generated for computing the ranking loss. 
\par
While $(p^{S1}_{j},p^{N1}_{j})$ is supposed to be the cross-border point pair, the edge point $p_{j}^{e}$ from noisy semantic labels are not reliable to represent the \emph{real} object border. In this section, we solve this problem by incorporating image gradient as another clue to detect real object borders, for the object border points usually possess the largest gradient value in the local area. We search the maximum image gradient point $p_{j}^{g}$ along the line $\mathbf{n}$, within a small distance range $r$ to the edge point $p_{j}^{e}$. As shown in Figure \ref{fig:edge_sample}, when sampling the cross-border point, we set $(p^{S1}_{j},p^{N1}_{j})$ to clamp both $p_{j}^{g}$ and $p_{j}^{e}$ to make the sampled points across both semantic and image gradient border. The margin pixel $\omega_{1}$ is set to $1$ for tight clamping. The distance range $r$ is set to $[-5,5]$ empirically, in order to strike a balance between finding the real borders and making the sampled pairs close to the borders.
For point $p^{S2}_{j}$ and $p^{N2}_{j}$, they are sampled besides the $p^{S1}_{j}$ and $p^{N1}_{j}$ respectively, within a distance range $\omega_{2}$ which is set to $[2.5,10]$. For each input image, we select all edge points and feed all corresponding quadruplets to compute the ranking loss in Equation \ref{eq:ranking_loss}.
Besides the cross border point quadruplets set $Q$, we also randomly sample a point pair set $O=\{(p^{1}_{k},p^{2}_{k})\}_{k=1}^{K}$, whose point pairs lie inside the semantic borders, to constrain the depth smoothness in a broader range.

\begin{table*}[th] 
\center
\scalebox{1.0}{
\begin{tabular}{|p{3.5cm}<{\raggedright}|c|c|c|c|c|c|c|c|}
\hline
\textbf{Method}        & Training & \cellcolor{my_blue}Abs Rel & \cellcolor{my_blue}Sq Rel & \cellcolor{my_blue}RMSE  & \cellcolor{my_blue}RMSE$_{log}$ & \cellcolor{my_red}$\delta < 1.25$ &\cellcolor{my_red} $\delta < 1.25^{2}$ & \cellcolor{my_red} $\delta < 1.25^{2}$ \\ \hline
Zhou \cite{zhou2017unsupervised}                   & M        & 0.183   & 1.595  & 6.709 & 0.270    & 0.734             & 0.902                 & 0.959                 \\
Mahjourian \cite{mahjourian2018unsupervised}             & M        & 0.163   & 1.240  & 6.220 & 0.250    & 0.762             & 0.916                 & 0.968                 \\
GeoNet \cite{yin2018geonet}                & M        & 0.155   & 1.296  & 5.857 & 0.233    & 0.793             & 0.931                 & 0.973                 \\
DDVO \cite{wang2018learning}                  & M        & 0.151   & 1.257  & 5.583 & 0.228    & 0.810             & 0.936                 & 0.974                 \\
CC \cite{ranjan2019competitive}              & M        & 0.140   & 1.070  & 5.326 & 0.217    & 0.826             & 0.941                 & 0.975                 \\
EPC++ \cite{luo2018every}              & M        & 1.029   & 1.070  & 5.350 & 0.216    & 0.816             & 0.941                 & 0.976                 \\
GLNet \cite{chen2019self}     & M        & 0.135   & 1.070  & 5.230 & 0.210    & 0.841             & 0.948                 & 0.980                 \\
Monodepth2 \cite{godard2019digging}             & M        & 0.115   & 0.903  & 4.863 & 0.193    & 0.877             & 0.959                 & 0.981                 \\
PackNet \cite{guizilini20203d}             & M        & 0.111   & \underline{0.785}  & \underline{4.601} & 0.189    & 0.878             & 0.960                 & {0.982}   \\
Johnston \cite{johnston2020self}             & M        & \underline{0.106}   & 0.861  & 4.699 & \underline{0.185}    & \underline{0.889}            & 0.962                 & {0.982} \\
\hline

Casser \cite{casser2019depth}& M+Inst        & 0.141   & 1.026  & 5.291 & 0.215    & 0.816             & 0.945                 & 0.979                 \\
Chen \cite{chen2019towards}             & M+Sem        & 0.118   & 0.905  & 5.096 & 0.211    & 0.839            & 0.945                 & 0.977 \\
Ochs \cite{ochs2019sdnet}             & D+Sem        & 0.116   & 0.945  & 4.916 & 0.208    & 0.861            & 0.952                 & 0.968 \\
Guizilini \cite{guizilini2020semantically} - PackNet             & M+Sem        & 0.102   & 0.698  & 4.381 & 0.178    & 0.896            & 0.964                 & 0.984 \\
Guizilini \cite{guizilini2020semantically} - Res50             & M+Sem        & 0.113   & 0.831  & 4.663 & 0.189    & 0.878            & \textbf{0.971}                 & \underline{0.983} \\

\textbf{Ours}          & M+Sem        & \textbf{0.103}   & \textbf{0.709}  & \textbf{4.471} & \textbf{0.180}    & \textbf{0.892}             & \underline{0.966}                 & \textbf{0.984}                 \\ \hline

Casser (+ref.) \cite{casser2019depth}& M+Inst        & 0.109   & 0.825  & 4.750 & 0.187    & 0.874             & 0.958                 & 0.983                 \\
GLNet (+ref.) \cite{chen2019self}     & M        & 0.099   & 0.796  & 4.743 & 0.186    & 0.884             & 0.955                 & 0.979                 \\
\textbf{Ours} (+ref.)          & M+Sem        & \textbf{0.095}   & \textbf{0.666}  & \textbf{4.252} & \textbf{0.172}    & \textbf{0.905}             & \textbf{0.968}                 & \textbf{0.984}                 \\ \hline
\end{tabular}
}
\vspace{-5pt}
\caption{\textbf{Quantitative results on KITTI 2015}. The best results are in \textbf{bold} and the second best results are \underline{underlined}. ``M'' refer to self-supervision methods using monocular images only. ``Inst'' and ``Sem'' denote methods which leverage instance or semantic segmentation information. ``-Res50'' refers to the method which uses Resnet-50 as the backbone encoder. ``+ref.'' represents using the online refinement operation. All results are reported within range $[0m,80m]$. The metrics marked in \textcolor{my_blue}{blue} mean ``lower is better'', while these in \textcolor{my_red}{red} refer to ``higher is better''. Our method outperforms the state-of-the-arts in most metrics by significant margins.}
\label{tab:kitti_cmp}
\vspace{-15pt}
\end{table*}

\vspace{-3pt}
\subsubsection{Semantically Uncertainty-aware Weighting} \label{sec:uncer_factor}
Although the cross-border sampling strategy in Section \ref{sec:pts_sampling} heuristically constrains the point pairs to cross the real object border, it will fail when the pre-computed segmentation mask deviates significantly from the groundtruth. To address this problem, we propose a semantically uncertainty-aware weighting factor $\gamma$ for Equation \ref{eq:ranking_loss}, which evaluate the quality of the input quadruplet $q_{j}$ by leveraging semantic probabilities of the cross-border point pair. The semantic probability map $\hat{S}_{b}$ is generated by the semantic branch $f_{S}$. Though not being the ground truth probability, it indeed reflects the semantic uncertainties across different areas. Thus when computing the ranking loss of the three point pairs inside $q_{j}$, the weighted loss $L^{Q}_{SR}(p^{a}_{j},p^{b}_{j})$ is 
\begin{equation}
   L^{Q}_{SR}(p^{a}_{j},p^{b}_{j})=\gamma \cdot L_{SR}(p^{a}_{j},p^{b}_{j}),
\end{equation}
where $p_{j}^{a},p_{j}^{b} \in q_{j}$, the weighting value $\gamma\in [\frac{1}{e},1]$ is decided by the differences of semantic probabilities between the point pair $p^{S1}_{j}$ and $p^{N1}_{j}$
\begin{equation}
   \gamma=\exp(-1/\max(\frac{\hat{S}_{b}(p^{S1}_{j})}{\hat{S}_{b}(p^{N1}_{j})},\frac{\hat{S}_{b}(p^{N1}_{j})}{\hat{S}_{b}(p^{S1}_{j})})).
\end{equation}
This means the more confident the semantic network is to its predicted cross-border points, the larger weight will be assigned to the three ranking losses. Note that for the points pairs $O$, $L^{O}_{SR}(p^{1}_{k},p^{2}_{k})=L_{SR}(p^{1}_{k},p^{2}_{k})$. The final ranking loss $L^{\prime}_{SR}$ can be formulated as

\begin{equation}
L^{\prime}_{SR}=\frac{1}{3J}\sum_{j}^{J}L^{Q}_{SR}(p^{a}_{j},p^{b}_{j}) + \frac{1}{K} \sum_{k}^{K} L^{O}_{SR}(p^{1}_{k},p^{2}_{k})
\end{equation}

\par
\subsection{Final Loss}
The final loss of the whole pipeline can be formulated as
\begin{equation}
L=L_{ph}+L_{M}+\delta_{s}L_{s}+\delta_{r}L^{\prime}_{SR},
\end{equation}
where $\delta_{s}$ is set to $0.001$ for common practice \cite{godard2019digging}, and $\delta_{r}$ is set to $0.001$ empirically that it strikes a balance between overall performance and depth sharpness/smoothness.

\begin{figure*}[htbp]
\centering
\subfigure{
\begin{minipage}[t]{0.245\linewidth}
    \centering
    \includegraphics[width=1\linewidth]{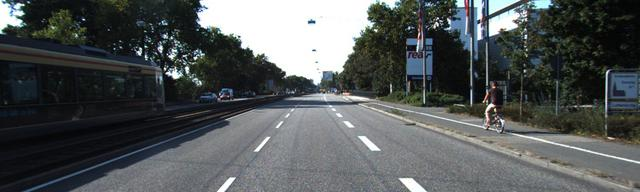} \\
    \vspace{1.5pt}
    \includegraphics[width=1\linewidth]{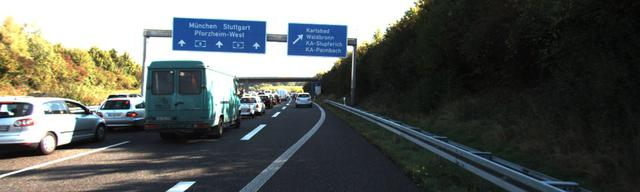} \\
    \vspace{1.5pt}
    \includegraphics[width=1\linewidth]{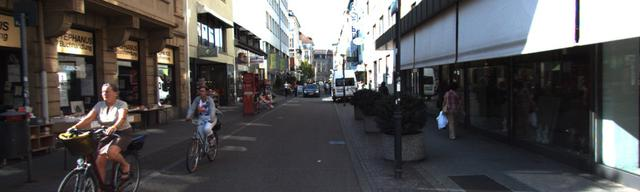} \\
    \vspace{1.5pt}
    \includegraphics[width=1\linewidth]{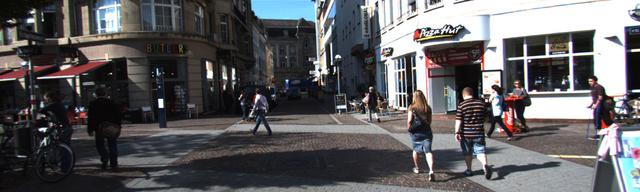} \\
    \vspace{1.5pt}
    \includegraphics[width=1\linewidth]{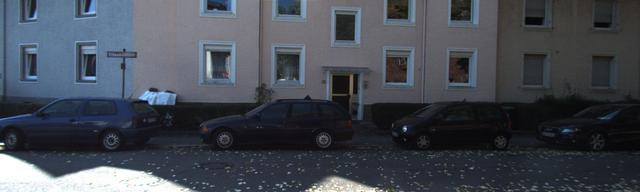} \\
    \vspace{1.5pt}
    \includegraphics[width=1\linewidth]{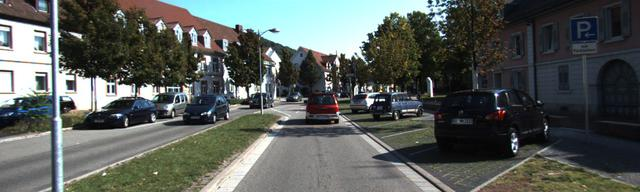} \\
    \vspace{1.5pt}
    \centerline{Input}
\end{minipage}%
\vspace{1.5pt}

\begin{minipage}[t]{0.245\linewidth}
    \centering
    \includegraphics[width=1\linewidth]{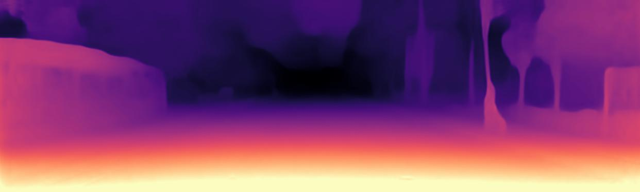} \\
    \vspace{1.5pt}
    \includegraphics[width=1\linewidth]{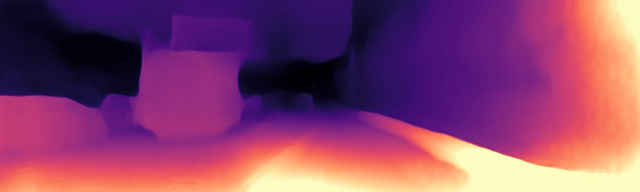} \\
    \vspace{1.5pt}
    \includegraphics[width=1\linewidth]{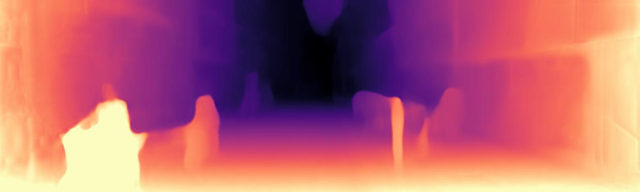} \\
    \vspace{1.5pt}
    \includegraphics[width=1\linewidth]{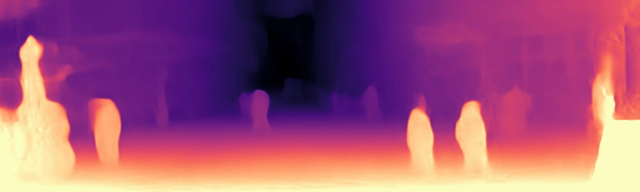} \\
    \vspace{1.5pt}
    \includegraphics[width=1\linewidth]{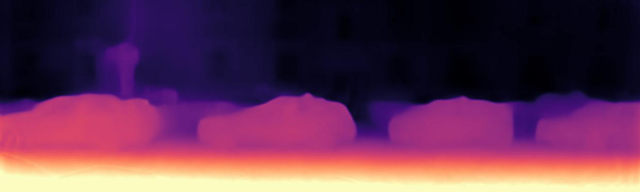} \\
    \vspace{1.5pt}
    \includegraphics[width=1\linewidth]{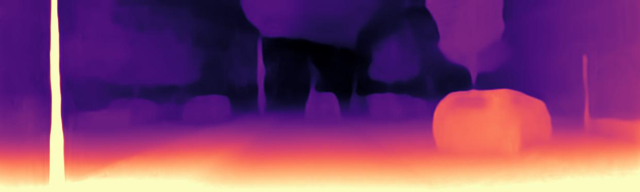} \\
    \vspace{1.5pt}
    \centerline{Monodepth2 \cite{godard2019digging}}
\end{minipage}%
\vspace{1.5pt}

\begin{minipage}[t]{0.245\linewidth}
    \centering
    \includegraphics[width=1\linewidth]{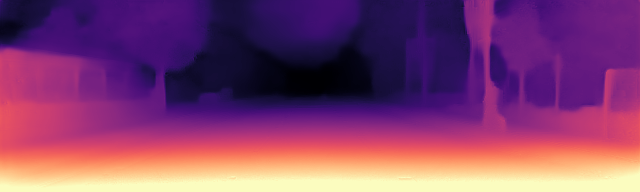} \\
    \vspace{1.5pt}
    \includegraphics[width=1\linewidth]{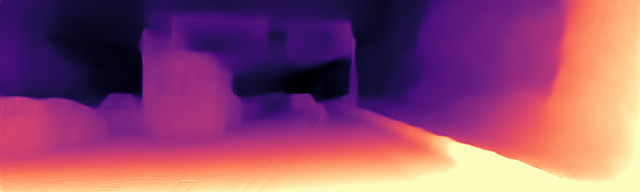} \\
    \vspace{1.5pt}
    \includegraphics[width=1\linewidth]{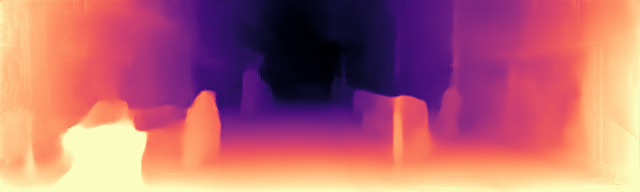} \\
    \vspace{1.5pt}
    \includegraphics[width=1\linewidth]{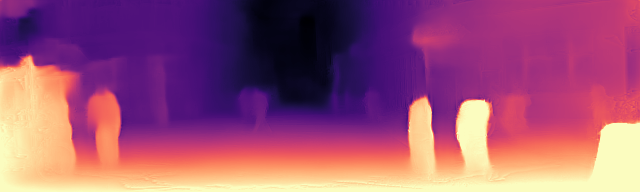} \\
    \vspace{1.5pt}
    \includegraphics[width=1\linewidth]{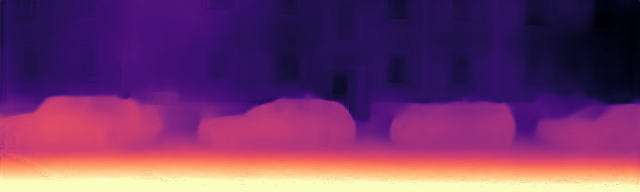} \\
    \vspace{1.5pt}
    \includegraphics[width=1\linewidth]{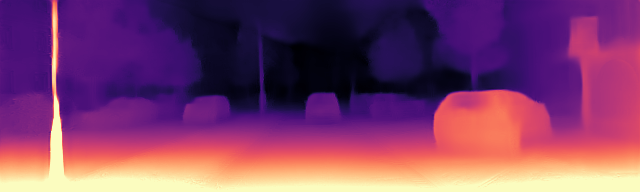} \\
    \vspace{1.5pt}
    \centerline{PackNet \cite{guizilini20203d}}
\end{minipage}%
\vspace{1.5pt}

\begin{minipage}[t]{0.245\linewidth}
    \centering
    \includegraphics[width=1\linewidth]{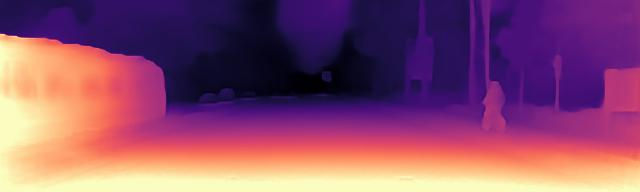} \\
    \vspace{1.5pt}
    \includegraphics[width=1\linewidth]{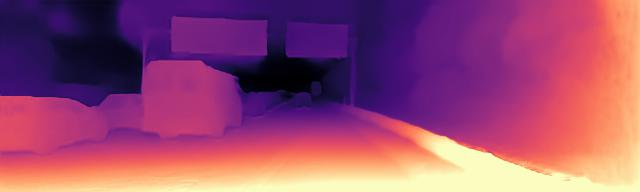}\\
    \vspace{1.5pt}
    \includegraphics[width=1\linewidth]{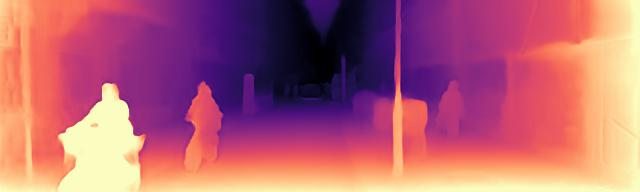} \\
    \vspace{1.5pt}
    \includegraphics[width=1\linewidth]{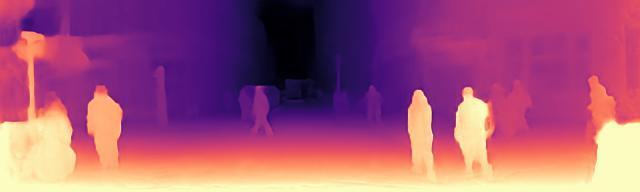} \\
    \vspace{1.5pt}
    \includegraphics[width=1\linewidth]{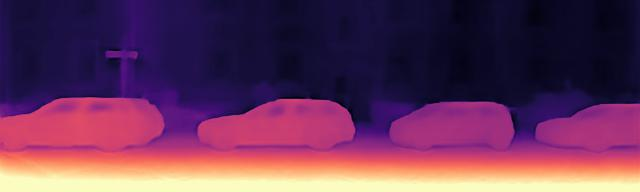} \\
    \vspace{1.5pt}
    \includegraphics[width=1\linewidth]{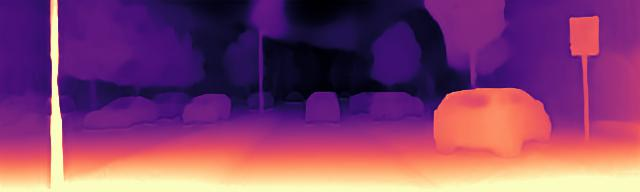} \\
    \vspace{1.5pt}
    \centerline{\textbf{Ours}}
\end{minipage}%
}
\centering
\vspace{-10pt}
\caption{\textbf{Qualitative comparison on KITTI}. Our method shows its superiority in producing better category-level depth predictions, and generates accurate depth predictions which are smooth inside the objects and sharp across object borders.}
\label{fig:viz_kitti}
\vspace{-15pt}
\end{figure*}

\section{Experiments}\label{sec:exp}
In this section, we conduct comprehensive comparisons to demonstrate the superiority of our method toward the state-of-the-arts, and validate the generalization ability of the proposed method in leveraging semantic information.
\subsection{Experimental Settings}
\textbf{Dataset.} We use the KITTI 2015 \cite{geiger2012we} dataset with Eigen's split \cite{eigen2014depth} and Zhou's \cite{zhou2017unsupervised} pre-preprocessing strategy, which leads to 39810 training images and 697 testing images. For the training of the semantic branch, we only use the pre-computed semantic maps from an off-the-shelf model \cite{zhu2019improving} for supervision. The model is pre-trained on Cityscape \cite{cordts2016cityscapes} (CS) and Mapillary Vistas dataset \cite{MVD2017} (V). We mainly use the model fine-tuned on 200 KITTI (K) labeled images to generate the dataset $\mathbf{Sem}_{CS+V+K}$. Consider the real world scenarios, we also maintain another dataset $\mathbf{Sem}_{CS+V}$ without fine-tunning on KITTI, to validate the feasibility of using the precomputed labels trained from cross-domain datasets. We follow the practice of \cite{zhu2020edge} to generate the binary semantic maps.
\par

\textbf{Implementation Details.}
We build our method on Monodepth2 \cite{godard2019digging} with ResNet-50 \cite{he2016deep} pre-trained on the ImageNet \cite{deng2009imagenet} as backbone. The model is trained on a single NVIDIA Tesla V100 GPU with batch size of 12. The learning rate is set to $10^{-4}$ and divided by $10$ for every $15$ epochs. After the network converges, we select the model of epoch 20 for testing. The input image size is set to $192 \times 640$ following \cite{godard2019digging}, and we compare our method with others on the same image resolution. We conduct online refinement following the practice of \cite{casser2019depth,chen2019self,shu2020feature} with batch size of $1$. The online refinement is performed $20$ iterations on each test image, there is no data augmentation during this process. During testing, the network generates depth from the input image, and the semantic label is used to guide the feature alignment of the depth and semantic branches.

\subsection{Results on KITTI}
We compare the proposed method with state-of-the-art monocular trained depth estimation methods on KITTI 2015 \cite{geiger2012we}, the quantitative results are shown in Table \ref{tab:kitti_cmp}. Note that the PackNet-based implementation of Guizilini \etal \cite{guizilini2020semantically} is $0.102$ in AbsRel. However, PackNet \cite{guizilini20203d} alone takes more than 120M parameters for training, which is not applicable under certain circumstances with limited computation resources. Thus, we select the general Resnet-50 as backbone, which is also the same as our method for fair comparison.
Our method outperforms state-of-the-art methods by a large margin on most evaluation metrics. The qualitative results are shown in Figure \ref{fig:viz_kitti}. The superiority of our method are illustrated in two aspects. Firstly, our method outperforms others in understanding depth from the category-level perspective. For instance, it successfully infers the correct depth on ``train'' and ``traffic sign'' area than other methods in the row $1\sim 2$ of Figure \ref{fig:viz_kitti}. Secondly, our method predicts high quality depth which is smooth inside object area (as the rider on row $3$), and is sharp across the object borders (as the borders of people, cars and traffic signs in row $4\sim 6$). More results can be found in the supplementary material.

\begin{table*}[th]
\center
\scalebox{0.950}{
\begin{tabular}{|p{3.9cm}<{\raggedright}|c|c|c|c|c|c|c|c|c|}
\hline
Method           & SSFA                     & SRL                      & \cellcolor{my_blue}Abs Rel & \cellcolor{my_blue}Sq Rel & \cellcolor{my_blue}RMSE  & \cellcolor{my_blue}RMSE$_{log}$ & \cellcolor{my_red}$\delta \le 1.25$ & $\cellcolor{my_red}\delta \le 1.25^{2}$ & $\cellcolor{my_red}\delta \le 1.25^{2}$ \\ \hline

Baseline         & \XSolidBrush    & \XSolidBrush    & 0.110   & 0.830  & 4.639 & 0.187        & 0.884             & 0.962                 & 0.982                 \\

Baseline + SSFA    & \checkmark & \XSolidBrush    & 0.107   & {0.777}  & 4.583 & 0.183        & 0.889             & 0.963                 & {0.983}                 \\

Baseline + SRL      & \XSolidBrush    & \checkmark & {0.107}   & 0.766  & {4.583} & {0.184}        & {0.887}             & {0.963}                 & {0.983}                 \\

Baseline + SSFA + SRL & \checkmark & \checkmark & \textbf{0.103}   & \textbf{0.709}  & \textbf{4.471} &\textbf{0.180}        & \textbf{0.892}             & \textbf{0.966}                 & \textbf{0.984}                 \\ \hline
\end{tabular}
}
\caption{\textbf{Ablation experiments}. We show the results of several ablated versions of our method on KITTI 2015 \cite{geiger2012we}. ``SSFA'' denotes the proposed SSFA scheme, and ``SRL'' refers to the semantic ranking loss. The best results are in \textbf{bold}.}
\label{tab:ablation}
\vspace{-15pt}
\end{table*}

\subsection{Ablation Study}
We evaluate the effectiveness of the proposed modules in Table \ref{tab:ablation} by comparing the different versions of our method. We see that the baseline model performs the worst among all, and our proposed contributions improve consistently upon the baseline. When combined together, both contributions lead to a significant improvement of the performance.

\subsection{Category-specific Depth Improvement}
\begin{figure}
\centering
\includegraphics[width=0.95\linewidth]{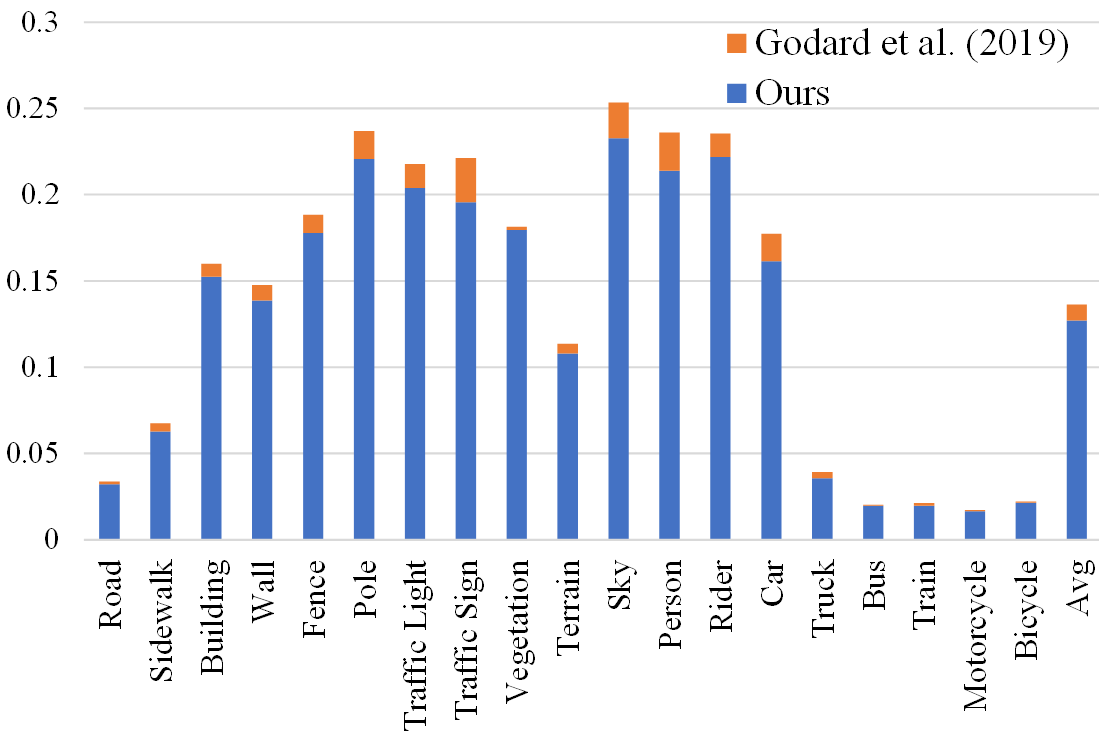}
\centering
\vspace{-5pt}
\caption{\textbf{Category-specific depth improvement.} We compare our method (blue) with the baseline (orange) using the metric of AbsRel. The rightmost is the average performance, which is the mean value of all categories' performance. Our method improves consistently across most categories, including the foreground (person, rider, traffic sign, \etc.) and the background (sky, fence, wall, \etc.) classes. 
}
\label{fig:cate_level_res}
\vspace{-5pt}
\end{figure}

To further analysis the category-level improvements, we evaluate the improvement of depth predictions with respect to their semantic labels. Due to the absence of semantic groundtruth, we use the fine-tuned predicted labels $\mathbf{Sem}_{CS+V+K}$ to specify the semantic area. We compare our method with the baseline model \cite{godard2019digging}, with AbsRel as the evaluation metric. As shown in Figure \ref{fig:cate_level_res}, our method shows improvements towards most of the categories, including not only the foreground categories (traffic signs, person, cars, \etc.), but also the background categories (sky, fence, wall, \etc.) which can not be directly seen from the binary mask. It indicates that the semantic features provided by the binary mask are capable to offer rich contextual information with the guidance of the SSFA.

\begin{figure}[htbp]
     \centering
     \subfigure{
         \begin{minipage}[t]{0.33\linewidth}
             \centering
             \includegraphics[width=1\linewidth]{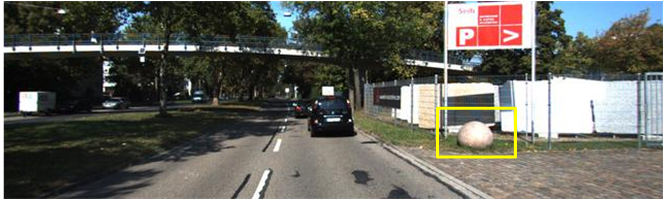} \\
             \vspace{0.02cm}
             \includegraphics[width=1\linewidth]{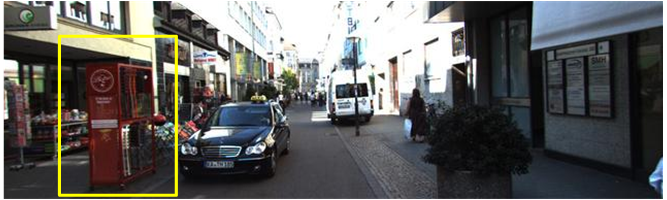} \\
             \vspace{0.02cm}
             \includegraphics[width=1\linewidth]{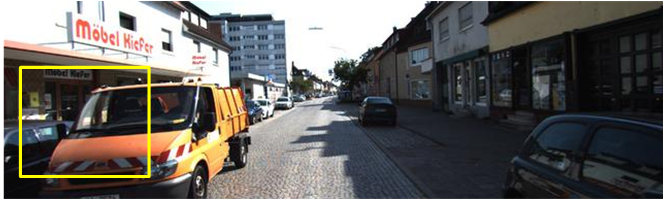} \\
             \vspace{0.02cm}
         \end{minipage}%
         \vspace{0.02cm}

         \begin{minipage}[t]{0.33\linewidth}
             \centering
             \includegraphics[width=1\linewidth]{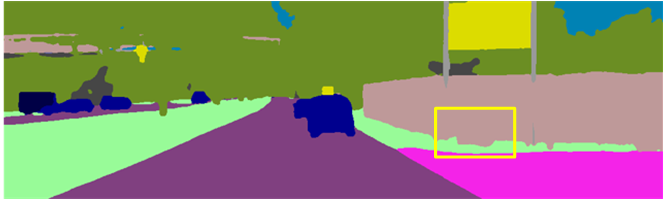} \\
             \vspace{0.02cm}
             \includegraphics[width=1\linewidth]{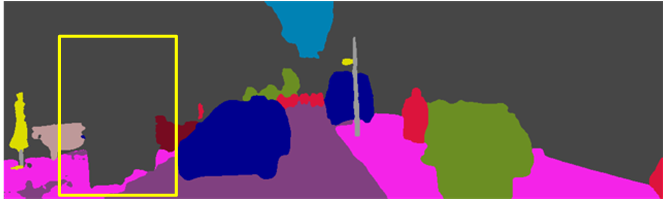} \\
             \vspace{0.02cm}
             \includegraphics[width=1\linewidth]{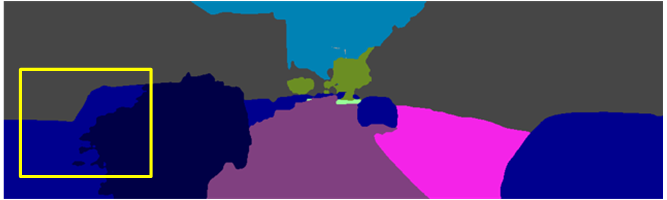}\\
             \vspace{0.02cm}
         \end{minipage}%
         \vspace{0.02cm}

         \begin{minipage}[t]{0.33\linewidth}
             \centering
             \includegraphics[width=1\linewidth]{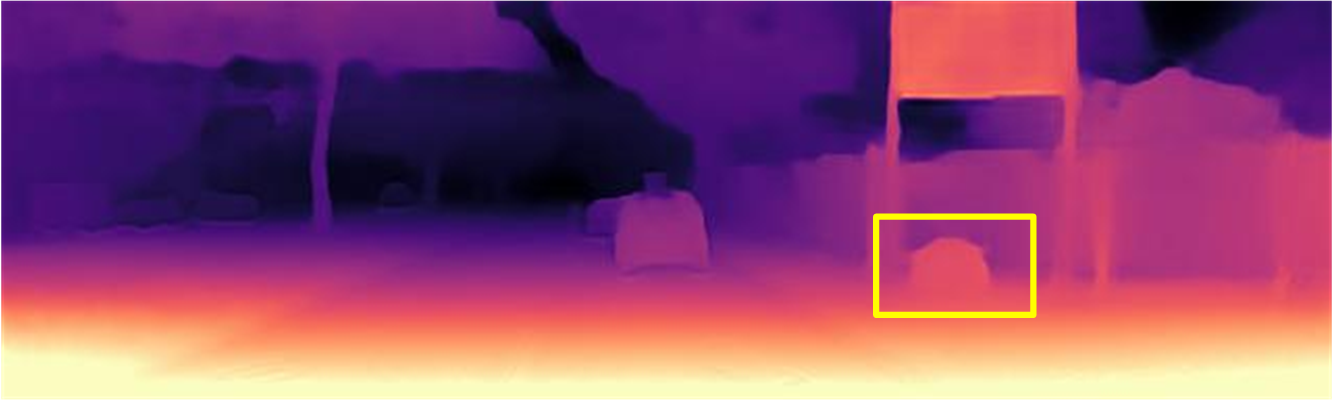} \\
             \vspace{0.02cm}
             \includegraphics[width=1\linewidth]{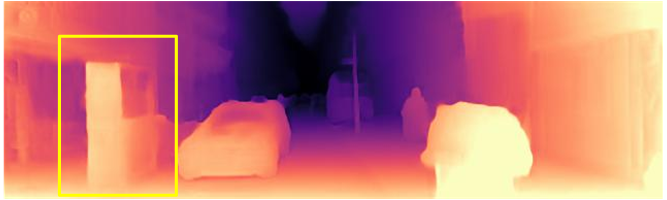} \\
             \vspace{0.02cm}
             \includegraphics[width=1\linewidth]{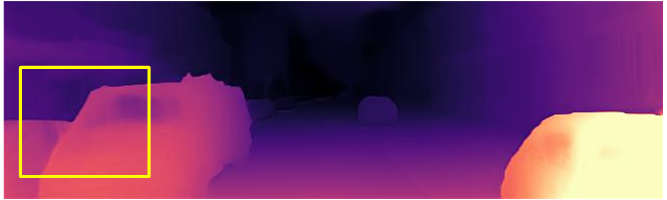} \\
             \vspace{0.02cm}
         \end{minipage}%
     }     
\centering
\vspace{-10pt}
\caption{\textbf{Depth from erroneous segmentation labels}. Despite different types of semantic degradation, our method still produce accurate depth maps.}
\label{fig:error_seg}
\vspace{-5pt}
\end{figure}

\subsection{Performance on Noisy Semantic Labels}
Due to the desperately lack of groundtruth semantic labels on real world scenarios, the noisy semantic inputs are inevitable during both training and testing phrase. We validate the effectiveness of our method in sampling point pairs across the real object borders (see Section \ref{sec:pts_sampling}). We use the pre-computed semantic label as guidance, and compare the proposed semantic-guided sampling strategy with the direct strategy which samples directly beside the edge points. We calculate the ratio of real cross-border points to all sampled pairs, using KITTI Semantics dataset \cite{geiger2012we} as groundtruth. 
The quantitative results are shown in Table \ref{tab:test_sempling}. Our sampling strategy improves upon the direct sampling strategy by a significant margin of $19.9\%$, which shows its great superiority in finding the real borders in the local area. 

\begin{table}
\begin{center}
\scalebox{0.95}{
\begin{tabular}{|p{3.0cm}<{\raggedright}|c|c|}
\hline
Sampling Strategy & Direct & The Proposed  \\ \hline
Accuracy (\%)      & 55.3   & \textbf{75.2} \\ \hline
\end{tabular}
}
\end{center}
\vspace{-10pt}
\caption{\textbf{The sampling accuracy of different strategies}. Our semantic-guided sampling strategy outperforms the direct method by a significant margin of $19.9 \% $.}
\label{tab:test_sempling}
\vspace{-18pt}
\end{table}

\par
During testing, we validate that our method can successfully handle erroneous semantic inputs as shown in Figure \ref{fig:error_seg}. For the input semantic labels, in the first row the stone is mistaken as the wall, in the second row the cabinet is not labeled and in the third row, the boundary of the left car is extremely imprecise. Despite the erroneous semantic predictions, our method still succeed in performing accurate depth predictions of the true scene structure. 

\begin{table}
\vspace{5pt}
\begin{center}
\scalebox{0.77}{
\begin{tabular}{|l|c|c|c|c|c|}
\hline
Dataset & mIoU$_{Full}$ & mIoU$_{Bi}$ & \cellcolor{my_blue}Abs Rel & \cellcolor{my_blue}RMSE & \cellcolor{my_red}$\delta \le 1.25$ \\ \hline
$\mathbf{Sem}_{CS+V}$   & 67.73                                & 93.54                               & 0.105                           & 4.516                       & 0.890                                     \\
$\mathbf{Sem}_{CS+V+K}$ & 79.90                                & 94.74                               & 0.103                           & 4.471                       & 0.892                                     \\ \hline
\end{tabular}
}
\end{center}
\vspace{-10pt}
\caption{\textbf{Performance on different datasets}. The binary segmentation IoU (mIoU$_{Bi}$) differs a little across different datasets, and our model achieves comparable results on cross-domain generated semantic labels.}
\label{tab:diff_semdata}
\vspace{-15pt}
\end{table}

\subsection{Study on Cross-domain Semantic Labels}\label{sec:diff_semdata}
Although fine-tuned only on $200$ KITTI semantic labels, the generated semantic dataset $\mathbf{Sem}_{CS+V+K}$ is still guided by groundtruth information.
To further validate the generalization ability, we train our model with semantic dataset $\mathbf{Sem}_{CS+V}$, which is acquired from totally cross-domain trained semantic model. As shown in Table \ref{tab:diff_semdata}, we found an interesting phenomenon that although the full segmentation mIoU (mIoU$_{Full}$) of $\mathbf{Sem}_{CS+V}$ is obviously worse than that of $\mathbf{Sem}_{CS+V+K}$, their binary mIoU (mIoU$_{Bi}$) is very close to each other ($93.54\%$ to $94.74\%$). It validates the advantage of the binary semantic maps, with which the performance of both semantic branch and semantic-guided sampling are less influenced by cross-domain predictions. At the meantime, our model trained on the cross-domain semantic dataset $\mathbf{Sem}_{CS+V}$ generalizes well, for its performance is only slightly behind with the one trained on $\mathbf{Sem}_{CS+V+K}$, while still outperforms the state-of-the-art methods \cite{guizilini20203d,godard2019digging,johnston2020self}.

\vspace{-2.5pt}
\section{Conclusion}
\vspace{-2.5pt}
In this paper, we improve self-supervised depth estimation via leveraging both implicit and explicit semantic guidance. We propose a semantic-guided spatial feature alignment scheme to implicitly model the semantic category-level information for depth estimation. And we propose a semantic-guided ranking loss to explicitly constrain the depth map's accuracy regarding to specific object categories. Extensive experiments show the superiority of the method. In future research, we plan to leverage the semantic information for depth refinement, to further improve the performance after the initial predictions.

{\small
\bibliographystyle{ieee_fullname}
\bibliography{egbib}
}

\end{document}


\title{Supplementary Material for: \\ ``Learning Depth via Leveraging Semantics: Self-supervised Monocular Depth Estimation with Both Implicit and Explicit Semantic Guidance''}


\author{First Author\\
Institution1\\
Institution1 address\\
{\tt\small firstauthor@i1.org}
\and
Second Author\\
Institution2\\
First line of institution2 address\\
{\tt\small secondauthor@i2.org}
}

\maketitle

\section*{Overview}
In this supplementary material, we show more implementation details as well as more experimental results.

\section{Network Details}
The network details are shown in Figure \ref{fig:sup_network}. We introduce the architectures of the depth decoder, semantic decoder as well as the SSFA module. Note that there are 5 SSFA modules on each layer of the decoding branch. In order to make the figure easier to read, we select the SSFA4 which is implemented on the first decoding layer (layer 4) as an example.

\begin{figure*}[ht]
\centering
\includegraphics[width=0.95\linewidth]{./figures/sup/network_arch.png}
\centering
\caption{\textbf{The network details of the method}. ``in\_chns'' and ``out\_chns'' refer to the number of input and output channels, ``resolution'' represents the downscaling factor with regard to the input image, and ``input'' stands for the input of each layer, where ``$\uparrow$'' means NN-based upsampling. In the depth decoder, ``SSFA'' denotes the proposed semantic-aware spatial feature alignment modules.}
\label{fig:sup_network}
\end{figure*}

\section{Parameter Selection for Semantic-guided Ranking Loss}
In this section, we discuss the parameter selection of the proposed semantic-guided ranking loss.

\subsection{Selection of the Distance Range $r$}
For the proposed cross-border point sampling strategy (Section 3.2.1 in the main paper), the distance range $r$ affects the performance of cross-border sampling. To find an appropriate value of $r$, we conduct quantitative experiment on KITTI Semantics \cite{geiger2012we}. We sample the cross-border point pairs under the guidance of pre-computed labels with varying values of $r$, and compute the inlier rate of the point pairs which lies across the real semantic borders. The results are shown in Table \ref{tab:r_value}. We can see that as $r$ is set from $[0,0]$ to $[-3,3]$, the inlier rate improves significantly from $55.30\%$ to $73.74\%$. However, when $r$ continues to change from $[-5,5]$ to $[-9,9]$, the inlier rate tends to grow slowly, and there is no obvious difference between inlier rates under $[-5,5]$ and $[-9,9]$. Consider that the closely sampled cross-border point pairs will bring more benefits to the removal of the depth ``bleeding effects'', we thus select $r$ to $[-5,5]$ to strike a balance between finding the real borders and making the two sampled points close to each other.

\renewcommand\arraystretch{1.2}
\begin{table}[h]
\centering
\begin{tabular}{|c|c|c|c|c|c|c|}
\hline
$r$          &$[0,0]$     & $[-1,1]$     & $[-3,3]$     & $[-5,5]$     & $[-7,7]$     & $[-9,9]$     \\ \hline
Inlier rate (\%) & 55.30 & 67.09 & 73.74 & 75.23 & 75.78 & 75.97 \\ \hline
\end{tabular}
\caption{\textbf{Cross-border inlier rate with different distance range}.}
\label{tab:r_value}
\vspace{-10pt}
\end{table}

\subsection{Sampling Strategy for the Random Point Pair Set $O$}
We illustrate in the main paper (Section 3.2.1) that we also randomly sample point pairs which lie inside the semantic object areas. First, $K^{\prime}$ point pairs are randomly sampled from the whole image, and we observe that setting $K^{\prime}=6000$ is sufficient to cover the image. The distance between the two points inside a pair is varying from $[-\varepsilon,\varepsilon]$ along both $\mathbf{x}$ and $\mathbf{y}$ axises. We set $\varepsilon=3$ that the depth smoothness constraint inside the object is only valid within the local area. Then, we locate the point pairs which lie inside the object areas with the pre-computed labels, yielding $K$ final point pairs to compute the depth loss. Note that the value of $K$ can be different across images, which is decided by the random sampling scheme as well as the different object areas.

\section{Further Discussions on the Semantic Branch}
In this section, we provide detailed information about the semantic branch, including foreground / background selection as well as the visual comparison between full and binary semantic maps.

\subsection{Foreground/Background Settings}
The selection of foreground and background categories is shown in Table \ref{tab:fbg}. Given the full semantic labels, we get the binary semantic map by assigning the foreground to $1$ and the background to $0$.

\begin{table}[h]
\centering
\begin{tabular}{|c|p{11.5cm}<{\raggedright}|}
\hline
\textbf{Foreground categories} & traffic sign, person, rider, car, truck, bus, train, motorcycle, bicycle, traffic light \\ \hline
\textbf{Background categories} & road, sidewalk, building, wall, fence, pole, vegetation, terrain, sky, others           \\ \hline
\end{tabular}
\caption{\textbf{Selection of foreground / background areas}.}
\label{tab:fbg}
\vspace{-10pt}
\end{table}

\subsection{Visual Comparison between Full and Binary Semantic Map}
We compare the semantic segmentation performance from model trained on $M_{CS+V+K}$ and $M_{CS+V}$ respectively on Figure \ref{fig:sem_labels}. We test the models on KITTI Semantics which has groundtruth labels. Row 3-4 are the error maps from full segmentation results, where the falsely segmented areas are marked with color. Row 5-6 are error maps from binary segmentation results, and the erroneous areas are marked with gray. We can see from the figure that the two models performs differently on the full semantic segmentation task with a remarkable margin, but when it comes to binary segmentation, the two models have similar performances. As illustrated in the main paper, this is caused by the fact that the mis-segmentation often occurs between categories belong solely to the foreground or the background. For instance, though the ``sidewalk'' is falsely segmented as ``road'' by model trained on $M_{CS+V}$ in the first column, the erroneous area does not show in the binary error map because both ``road'' and ``sidewalk'' belong to the background category. This validate the advantage of using the binary semantic label that it is less influenced by the general semantic segmentation outputs.

\begin{figure}[ht]
\centering
\subfigure{
 \begin{minipage}[c]{0.01\linewidth}
            \centering
             \small
            \rotatebox{90}{\makebox{Input}}
  \end{minipage}
  \begin{minipage}[c]{1\linewidth}
      \centering
      \includegraphics[width=0.23\linewidth]{./figures/sup/image_2/000024_10.png}
      \includegraphics[width=0.23\linewidth]{./figures/sup/image_2/000030_10.png}
      \includegraphics[width=0.23\linewidth]{./figures/sup/image_2/000102_10.png}
      \includegraphics[width=0.23\linewidth]{./figures/sup/image_2/000130_10.png}
  \end{minipage}%
}\vspace{-4pt}
\subfigure{
  \begin{minipage}[c]{0.01\linewidth}
            \centering
             \small
            \rotatebox{90}{\makebox{\shortstack[c]{Semantic\\GT}}}
  \end{minipage}
  \begin{minipage}[c]{1\linewidth}
      \centering
      \includegraphics[width=0.23\linewidth]{./figures/sup/semantic_rgb/000024_10.png}
      \includegraphics[width=0.23\linewidth]{./figures/sup/semantic_rgb/000030_10.png}
      \includegraphics[width=0.23\linewidth]{./figures/sup/semantic_rgb/000102_10.png}
      \includegraphics[width=0.23\linewidth]{./figures/sup/semantic_rgb/000130_10.png}
  \end{minipage}%
}\vspace{-8pt}
\subfigure{
 \begin{minipage}[c]{0.01\linewidth}
            \centering
             \small
            \rotatebox{90}{\makebox{\shortstack[c]{Full\\$E_{CS+V+K}$}}}
  \end{minipage}
  \begin{minipage}[c]{1\linewidth}
      \centering
      \includegraphics[width=0.23\linewidth]{./figures/sup/error_c+v+k_19/000024_10.png}
      \includegraphics[width=0.23\linewidth]{./figures/sup/error_c+v+k_19/000030_10.png}
      \includegraphics[width=0.23\linewidth]{./figures/sup/error_c+v+k_19/000102_10.png}
      \includegraphics[width=0.23\linewidth]{./figures/sup/error_c+v+k_19/000130_10.png}
  \end{minipage}%
}\vspace{-8pt}
\subfigure{
 \begin{minipage}[c]{0.01\linewidth}
            \centering
             \small
            \rotatebox{90}{\makebox{\shortstack[c]{Full\\$E_{CS+V}$}}}
  \end{minipage}
  \begin{minipage}[c]{1\linewidth}
      \centering
      \includegraphics[width=0.23\linewidth]{./figures/sup/error_c+v_19/000024_10.png}
      \includegraphics[width=0.23\linewidth]{./figures/sup/error_c+v_19/000030_10.png}
      \includegraphics[width=0.23\linewidth]{./figures/sup/error_c+v_19/000102_10.png}
      \includegraphics[width=0.23\linewidth]{./figures/sup/error_c+v_19/000130_10.png}
  \end{minipage}%
}\vspace{-8pt}
\subfigure{
 \begin{minipage}[c]{0.01\linewidth}
            \centering
             \small
            \rotatebox{90}{\makebox{\shortstack[c]{Binary\\$E_{CS+V+K}$}}}
  \end{minipage}
  \begin{minipage}[c]{1\linewidth}
      \centering
      \includegraphics[width=0.23\linewidth]{./figures/sup/error_c+v+k_2/000024_10.png}
      \includegraphics[width=0.23\linewidth]{./figures/sup/error_c+v+k_2/000030_10.png}
      \includegraphics[width=0.23\linewidth]{./figures/sup/error_c+v+k_2/000102_10.png}
      \includegraphics[width=0.23\linewidth]{./figures/sup/error_c+v+k_2/000130_10.png}
  \end{minipage}%
}\vspace{-8pt}
\subfigure{
 \begin{minipage}[c]{0.01\linewidth}
            \centering
             \small
            \rotatebox{90}{\makebox{\shortstack[c]{Binary\\$E_{CS+V}$}}}
  \end{minipage}
  \begin{minipage}[c]{1\linewidth}
      \centering
      \includegraphics[width=0.23\linewidth]{./figures/sup/error_c+v_2/000024_10.png}
      \includegraphics[width=0.23\linewidth]{./figures/sup/error_c+v_2/000030_10.png}
      \includegraphics[width=0.23\linewidth]{./figures/sup/error_c+v_2/000102_10.png}
      \includegraphics[width=0.23\linewidth]{./figures/sup/error_c+v_2/000130_10.png}
  \end{minipage}%
}

\centering
\caption{\textbf{Example of semantic labels}. ``$E$'' refers to the semantic error maps where the falsely segmented areas are marked with color. ``Full'' denotes the full semantic segmentation results and ``Binary'' stands for the binary segmentation results. ``$CS+V+K$'' means the segmentation model is trained with dataset $M_{CS+V+K}$, and ``$CS+V$'' means the model trained with $M_{CS+V+K}$.}
\label{fig:sem_labels}
\end{figure}

\subsection{Evaluation on the Semantic Branch}
We perform quantitative comparison between our semantic branch and other state-of-the-art segmentation methods in terms of binary segmentation accuracy. We use KITTI Semantics dataset \cite{geiger2012we} for evaluation and we use the $M_{CS+V}$-generated semantic labels to train our model. We select the evaluation metric of binary mIoU and DICE. The results are shown in Table \ref{tab:sem_prediction}. As presented in the last two columns, our semantic branch learns well from the supervisory pseudo semantic labels. At the mean time, our model achieves comparable or even better performance toward current state-of-the-art semantic segmentation methods \cite{li2020improving,li2020gated,li2020semantic} in terms of binary segmentation accuracy, which demonstrates the effectiveness of the proposed semantic architecture.

\begin{table}[h]
\center
\scalebox{1.0}{
\begin{tabular}{|p{2.2cm}<{\raggedright}|c|c|c||c|c|}
\hline
Method      &  SFNet \cite{li2020semantic}    & GFFNet \cite{li2020gated} & DeSegNet\cite{li2020improving} & \textbf{Ours} &  $M_{CS+V}$\\ \hline
mIoU$_{Bi}$ (\%) & 89.23 &92.43&  92.94  &     91.60  & 93.54     \\ \hline
DICE (\%) & 95.95 &  94.07 &  96.58  &  95.48  &   96.24   \\ \hline
\end{tabular}
}
\caption{\textbf{Binary semantic segmentation accuracy}. We compare the binary prediction accuracy among our semantic branch, supervisory pseudo labels and other semantic segmentation methods.}
\label{tab:sem_prediction}
\end{table}

\section{Extra Results for KITTI}
In this section, we provide extra quantitative and qualitative results to further illustrate the superiority of our method.

\subsection{Quantitative Results on Improved KITTI}
We provide quantitative comparison on the Improved KITTI dataset \cite{geiger2012we}. As shown in Table \ref{tab:app_improved_kitti}, our method improves upon previous methods and produces competitive or even better results towards the state-of-the-art \cite{guizilini20203d}. When implemented with the online refinement strategy, our method further outperforms all the methods, which shows its effectiveness toward real-world scenarios.

\begin{table*}[t] 
\center
\scalebox{1.0}{
\begin{tabular}{|p{3.5cm}<{\raggedright}|c|c|c|c|c|c|c|c|}
\hline
\textbf{Method}        & Training & \cellcolor{my_blue}Abs Rel & \cellcolor{my_blue}Sq Rel & \cellcolor{my_blue}RMSE  & \cellcolor{my_blue}RMSE$_{log}$ & \cellcolor{my_red}$\delta < 1.25$ &\cellcolor{my_red} $\delta < 1.25^{2}$ & \cellcolor{my_red} $\delta < 1.25^{2}$ \\ \hline

Monodepth \cite{godard2017unsupervised}              & S        & 0.109  & 0.811  & 4.568 &0.166    &0.877             & 0.967                & 0.988               \\
SuperDepth \cite{pillai2019superdepth} +pp        & S        & 0.090 & 0.542  & 3.967 & 0.144    &0.901           & 0.976                 & 0.993                 \\ 
Monodepth2 \cite{godard2019digging}             & S        & 0.085   & 0.537 &3.868 &0.139    & 0.912             &0.979                & 0.993                \\
\hline
EPC++\cite{luo2018every}            & MS        & 0.123  & 0.754  &4.453 &0.172   & 0.863             & 0.964                & 0.989                 \\
Monodepth2 \cite{godard2019digging}             & MS        &{0.080}    & {0.466} &{3.681} &{0.127}    & {0.926}             &{0.985}                & {0.995}                \\
\hline
Zhou \cite{zhou2017unsupervised}                   & M        & 0.176   & 1.532 & 6.129 &0.244    & 0.758            & 0.921                 & 0.971                 \\
Mahjourian \cite{mahjourian2018unsupervised}             & M        & 0.134  & 0.983  &5.501& 0.203    & 0.827             & 0.944                 & 0.981                 \\
GeoNet \cite{yin2018geonet}                & M        & 0.132   & 0.994  & 5.240 &0.193    &0.833            & 0.953                 & 0.985                 \\
DDVO \cite{wang2018learning}                  & M        & 0.126   & 0.866  & 4.932 &0.185    & 0.851             & 0.958                 & 0.986                 \\
CC \cite{ranjan2019competitive}              & M        & 0.123   &0.881 & 4.834 & 0.181    & 0.860             & 0.959                 & 0.985                \\
EPC++ \cite{luo2018every}              & M        & 0.120   & 0.789  &4.755 &0.177    & 0.856             & 0.961                 & 0.987                 \\
Monodepth2 \cite{godard2019digging}             & M        & 0.090   & 0.545  & 3.942 & 0.137    & 0.914             &0.983                & 0.995                \\
PackNet \cite{guizilini20203d}          & M        & {0.078}   & {0.420}  & {3.485} & {0.121}    & {0.931}             &{0.986}                 & {0.996}                 \\ 

\textbf{Ours}          & M        & {0.081}   & {0.418}  & {3.564} & {0.123}    & {0.927}             &{0.987}                 & {0.997}                 \\ 
\textbf{Ours} (-ref.)          & M        & \textbf{0.070}   & \textbf{0.353}  & \textbf{3.252} & \textbf{0.110}    & \textbf{0.943}             &\textbf{0.990}                 & \textbf{0.997} \\
\hline
\end{tabular}
}
\caption{\textbf{Quantitative results on KITTI improved ground truth}.
``S'' and ``M'' refer to self-supervision methods trained using stereo images and monocular images, respectively. Results are presented without any post-processing. Best results are in \textbf{bold}. The metrics marked in \textcolor{my_blue}{blue} mean ``lower is better'', while these in \textcolor{my_red}{red} refer to ``higher is better''. Our method still produce competitive or better results compared to the state-of-the-art methods.}
\label{tab:app_improved_kitti}
\end{table*}

\subsection{More Qualitative Results}
\begin{figure*}[htbp]
\centering
\subfigure{
    \begin{minipage}[t]{0.23\linewidth}
        \centering
        \includegraphics[width=1\linewidth]{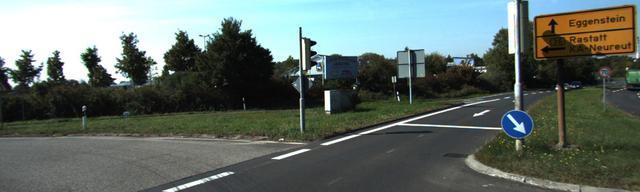} \\
        \vspace{0.3cm}
        \includegraphics[width=1\linewidth]{./figures/sup/sup_kitti/3_raw.png} \\
        \vspace{0.3cm}
        \includegraphics[width=1\linewidth]{./figures/sup/sup_kitti/21_raw.png} \\
        \vspace{0.3cm}
        \includegraphics[width=1\linewidth]{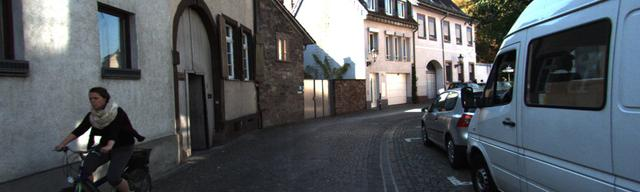} \\
        \vspace{0.3cm}
        \includegraphics[width=1\linewidth]{./figures/sup/sup_kitti/7_raw.png} \\
        \vspace{0.3cm}
        \includegraphics[width=1\linewidth]{./figures/sup/sup_kitti/17_raw.png} \\
        \vspace{0.3cm}
       \includegraphics[width=1\linewidth]{./figures/main/5_raw.png} \\
	    \vspace{0.3cm}
        \includegraphics[width=1\linewidth]{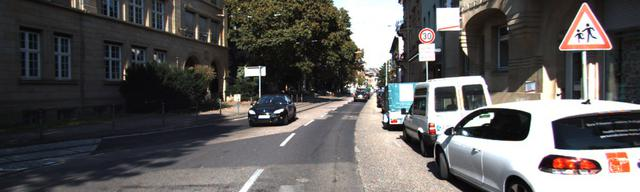} \\
	    \vspace{0.3cm}
        \includegraphics[width=1\linewidth]{./figures/sup/sup_kitti/19_raw.png} \\
        \vspace{0.3cm}
        \includegraphics[width=1\linewidth]{./figures/sup/sup_kitti/18_raw.png} \\
        \vspace{0.3cm}
        \includegraphics[width=1\linewidth]{./figures/sup/sup_kitti/20_raw.png} \\
        \vspace{0.3cm}
        \includegraphics[width=1\linewidth]{./figures/sup/sup_kitti/16_raw.png} \\
        \vspace{0.1cm}
        
        \centerline{Input}
    \end{minipage}%
    \vspace{0.02cm}

        \begin{minipage}[t]{0.23\linewidth}
        \centering
        \includegraphics[width=1\linewidth]{./figures/sup/sup_kitti/15_mono2.png} \\
        \vspace{0.3cm}
        \includegraphics[width=1\linewidth]{./figures/sup/sup_kitti/3_mono2.png} \\
        \vspace{0.3cm}
        \includegraphics[width=1\linewidth]{./figures/sup/sup_kitti/21_mono2.png} \\
        \vspace{0.3cm}
        \includegraphics[width=1\linewidth]{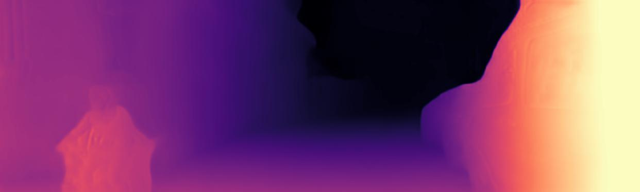} \\
        \vspace{0.3cm}
        \includegraphics[width=1\linewidth]{./figures/sup/sup_kitti/7_mono2.png} \\
        \vspace{0.3cm}
        \includegraphics[width=1\linewidth]{./figures/sup/sup_kitti/17_mono2.png} \\
        \vspace{0.3cm}
        \includegraphics[width=1\linewidth]{./figures/main/5_mono2.png} \\
	    \vspace{0.3cm}
        \includegraphics[width=1\linewidth]{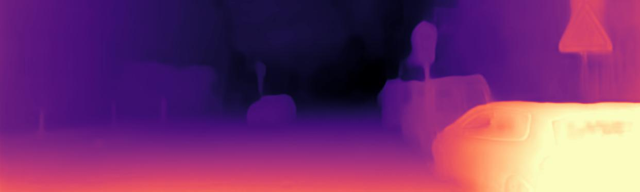} \\
	    \vspace{0.3cm}
        \includegraphics[width=1\linewidth]{./figures/sup/sup_kitti/19_mono2.png} \\
        \vspace{0.3cm}
        \includegraphics[width=1\linewidth]{./figures/sup/sup_kitti/18_mono2.png} \\
        \vspace{0.3cm}
        \includegraphics[width=1\linewidth]{./figures/sup/sup_kitti/20_mono2.png} \\
        \vspace{0.3cm}
        \includegraphics[width=1\linewidth]{./figures/sup/sup_kitti/16_mono2.png} \\
        \vspace{0.1cm}
        \centerline{Monodepth2 \cite{godard2019digging}}
    \end{minipage}%
    \vspace{0.02cm}

    \begin{minipage}[t]{0.23\linewidth}
        \centering
        \includegraphics[width=1\linewidth]{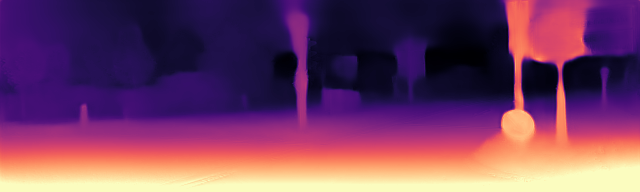} \\
        \vspace{0.3cm}
        \includegraphics[width=1\linewidth]{./figures/sup/sup_kitti/3_pack.png} \\
        \vspace{0.3cm}
        \includegraphics[width=1\linewidth]{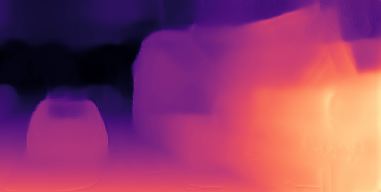} \\
        \vspace{0.3cm}
        \includegraphics[width=1\linewidth]{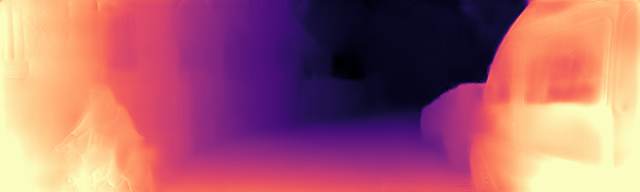} \\
        \vspace{0.3cm}
        \includegraphics[width=1\linewidth]{./figures/sup/sup_kitti/7_pack.png} \\
        \vspace{0.3cm}
        \includegraphics[width=1\linewidth]{./figures/sup/sup_kitti/17_pack.png} \\
        \vspace{0.3cm}
        \includegraphics[width=1\linewidth]{./figures/main/5_pack.png} \\
	    \vspace{0.3cm}
        \includegraphics[width=1\linewidth]{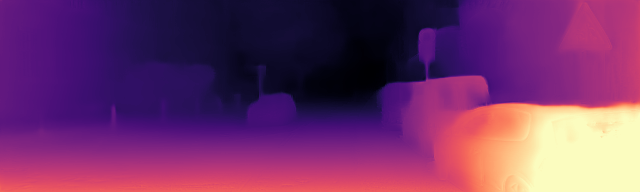} \\
	    \vspace{0.3cm}
        \includegraphics[width=1\linewidth]{./figures/sup/sup_kitti/19_pack.png} \\
        \vspace{0.3cm}
        \includegraphics[width=1\linewidth]{./figures/sup/sup_kitti/18_pack.png} \\
        \vspace{0.3cm}
        \includegraphics[width=1\linewidth]{./figures/sup/sup_kitti/20_pack.png} \\
        \vspace{0.3cm}
        \includegraphics[width=1\linewidth]{./figures/sup/sup_kitti/16_pack.png} \\
        \vspace{0.1cm}
        \centerline{PackNet \cite{guizilini20203d}}
    \end{minipage}%
    \vspace{0.02cm}

    \begin{minipage}[t]{0.23\linewidth}
        \centering
        \includegraphics[width=1\linewidth]{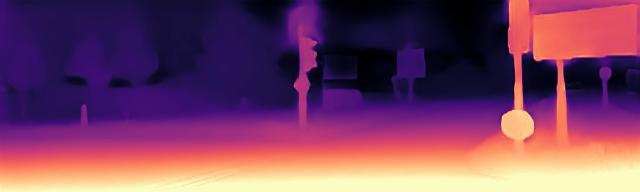} \\
        \vspace{0.3cm}
        \includegraphics[width=1\linewidth]{./figures/sup/sup_kitti/3_ours.png} \\
        \vspace{0.3cm}
        \includegraphics[width=1\linewidth]{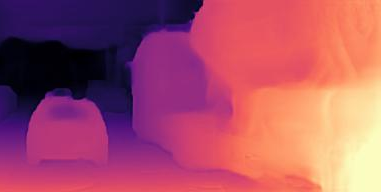} \\
        \vspace{0.3cm}
        \includegraphics[width=1\linewidth]{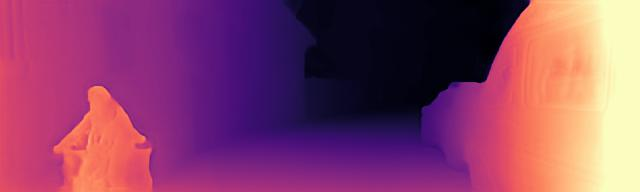} \\
        \vspace{0.3cm}
        \includegraphics[width=1\linewidth]{./figures/sup/sup_kitti/7_ours.png} \\
        \vspace{0.3cm}
        \includegraphics[width=1\linewidth]{./figures/sup/sup_kitti/17_ours.png} \\
        \vspace{0.3cm}
        \includegraphics[width=1\linewidth]{./figures/main/5_ours.png} \\
	    \vspace{0.3cm}
        \includegraphics[width=1\linewidth]{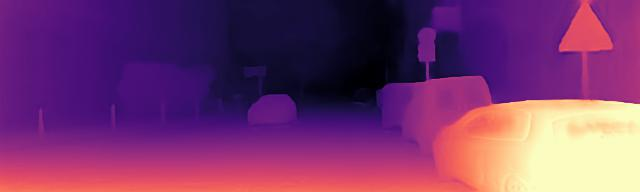} \\
	    \vspace{0.3cm}
        \includegraphics[width=1\linewidth]{./figures/sup/sup_kitti/19_ours.png} \\
        \vspace{0.3cm}
        \includegraphics[width=1\linewidth]{./figures/sup/sup_kitti/18_ours.png} \\
        \vspace{0.3cm}
        \includegraphics[width=1\linewidth]{./figures/sup/sup_kitti/20_ours.png} \\
        \vspace{0.3cm}
        \includegraphics[width=1\linewidth]{./figures/sup/sup_kitti/16_ours.png} \\
        \vspace{0.1cm}
        \centerline{\textbf{Ours}}
    \end{minipage}%

}

\centering
\caption{\textbf{Additional qualitative comparisons on KITTI dataset}. All methods are trained on KITTI with Eigen's \cite{eigen2014depth} split.}
\label{fig:sup_kitti}
\vspace{-8pt}
\end{figure*}

We provide more qualitative comparisons on KITTI dataset. As shown in Figure \ref{fig:sup_kitti}, our model outperforms consistently toward the state-of-the-art methods. It not only conducts consistent depth estimation (smooth depth inside objects and sharp depth across the borders) with the semantic objects, but also produces globally accurate estimations which conform to the semantic contextual information.


{\small
\bibliographystyle{ieee_fullname}
\bibliography{egbib}
}